%% file: main.tex
\pdfoutput=1

\documentclass[11pt]{article}

\usepackage{acl}

\usepackage{times}
\usepackage{latexsym}

\usepackage[T1]{fontenc}

\usepackage[utf8]{inputenc}

\usepackage{microtype}

\usepackage{caption}
\usepackage{subcaption}
\usepackage[ruled,vlined]{algorithm2e}
\usepackage{smile}

\newcommand{\ty}{\tilde{\mathbf{y}}}
\newcommand{\dd}{\mathrm{d}}
\newcommand{\ours}{{\sf DRIFT}}

\title{Self-Training with Differentiable Teacher}

\author{Simiao Zuo$^\diamond$\thanks{\hspace{0.03in} Equal contribution. Corresponding authors.}, \
Yue Yu$^{\diamond *}$, \ Chen Liang$^\diamond$, \ 
Haoming Jiang$^\Box$\thanks{\hspace{0.03in} Work was done at Georgia Institute of Technology.}, \
Siawpeng Er$^\diamond$, \\
\textbf{Chao Zhang$^\diamond$, \ Tuo Zhao$^\diamond$ and Hongyuan Zha$^\ddagger$} \\
$^\diamond$Georgia Institute of Technology \ \ $^\Box$Amazon \\
$^\ddagger$The Chinese University of Hong Kong, Shenzhen \\
\texttt{\{simiaozuo,yueyu,cliang73,ser8,chaozhang,tourzhao\}@gatech.edu}\\
\texttt{jhaoming@amazon.com} \quad \texttt{zhahy@cuhk.edu.cn}
}

\begin{document}
\maketitle

\begin{abstract}
Self-training achieves enormous success in various semi-supervised and weakly-supervised learning tasks. The method can be interpreted as a teacher-student framework, where the teacher generates pseudo-labels, and the student makes predictions. The two models are updated alternatingly.
However, such a straightforward alternating update rule leads to training instability.
This is because a small change in the teacher may result in a significant change in the student.
To address this issue, we propose {\ours}, short for differentiable self-training, that treats teacher-student as a Stackelberg game. In this game, a leader is always in a more advantageous position than a follower.
In self-training, the student contributes to the prediction performance, and the teacher controls the training process by generating pseudo-labels. Therefore, we treat the student as the leader and the teacher as the follower.
The leader procures its advantage by acknowledging the follower's strategy, which involves differentiable pseudo-labels and differentiable sample weights.
Consequently, the leader-follower interaction can be effectively captured via Stackelberg gradient, obtained by differentiating the follower's strategy.
Experimental results on semi- and weakly-supervised classification and named entity recognition tasks show that our model outperforms existing approaches by large margins.
\end{abstract}

\input{0-introduction}
\input{0-background}
\input{0-model}

\input{0-experiments}
\input{0-conclusion}

\input{0-acknowledgement}
\input{0-broader-impact}

\bibliography{anthology,custom}
\bibliographystyle{acl_natbib}

\clearpage 
\appendix
\input{0-appendix}

\end{document}

%% file: 0-introduction.tex
\section{Introduction}

Self-training is a classic method that was first proposed for semi-supervised learning~\cite{rosenberg2005semi, lee2013pseudo}.
It is also interpreted as a regularization method~\cite{mobahi2020self}, and is extended to weakly-supervised learning and domain adaptation~\cite{meng2018weakly}.
The approach has gain popularity in many applications. For example, in conjunction with pre-trained language models~\cite{devlin2018bert}, self-training has demonstrated superior performance on tasks such as natural language understanding~\cite{du2020self}, named entity recognition~\cite{liang2020bond}, and question answering~\cite{sachan2018self}. 

Conventional self-training can be interpreted as a teacher-student framework. Within this framework, a teacher model generates pseudo-labels for the unlabeled data. Then, a student model updates its parameters by minimizing the discrepancy between its predictions and the pseudo-labels.
The teacher subsequently refines its parameters based on the updated version of the student using pre-defined rules. Such rules include minimizing a loss function~\cite{pham2020meta}, copying the student's parameters~\cite{rasmus2015semi}, and integrating models from previous iterations~\citep{laine2016temporal, tarvainen2017mean}.
The above procedures are operated iteratively.

Computationally, the alternating update procedure often causes training instability.
Such instability comes from undesired interactions between the teacher and the student.
In practice, we often use stochastic gradient descent to optimize the student, and the noise of the stochastic gradient can cause oscillation during training.
This means in a certain iteration, the student is optimized towards a certain direction; while in the next iteration, it may be optimized toward a drastically different direction.
Such a scenario renders the optimization ill-conditioned.
Moreover, the student model's gradient is determined by the pseudo-labels generated by the teacher.
Because of the training instability, a small change in the pseudo-labels may result in a substantial change in the student.


To resolve this issue, we propose {\ours} (\textbf{d}iffe\textbf{r}ent\textbf{i}able sel\textbf{f}-\textbf{t}raining), where we formulate self-training as a Stackelberg game~\cite{von2010market}. The concept arises from economics, where there are two players, called the leader and the follower. In a Stackelberg game, the leader is always in an advantageous position by acknowledging the follower's strategy.
Within the self-training framework, we grant the student a higher priority than the teacher. This is because the teacher serves the purpose of generating intermediate pseudo-labels, such that the student can behave well on the task.
The student (i.e., the leader) procures its advantage by considering what the response of the teacher (i.e., the follower) will be, i.e., how will the follower react after observing the leader's move. Then, the leader makes its move, in anticipation of the predicted response of the follower.
We remark that the Stackelberg game formulation has also been used in other domains such as adversarial training \citep{zuo2021adversarial}.

We highlight that in {\ours}, the student has a higher priority than the teacher. In contrast, in conventional self-training, the two models are treated equally and have the same priority.
When using conventional self-training, the student only reacts to what the teacher has generated. In differentiable self-training, the student recognizes the teacher's strategy and reacts to what the teacher is anticipated to response.
In this way, we can find a better descent direction for the student, such that training can be stabilized.

To facilitate the leader's advantage, our framework treats the follower's strategy (i.e., pseudo-labels generated by the teacher) as a function of the leader's decision (i.e., the student's parameters). In this way, differentiable self-training can be viewed solely as a function of the student's parameters. Therefore, the problem can be efficiently solved using gradient descent.

Besides pseudo-labels, the teacher can also generate sample weights~\cite{freund1997decision, kumar2010self, malisiewicz2011ensemble}. Sample reweighting associates low-confidence samples with small weights, such that the influence of noisy labels can be effectively reduced.
Similar to pseudo-labels, sample weights and the student model are also updated iteratively.
As such, we can further equip {\ours} with differentiable sample weights. This can be achieved by integrating the weights as a part of the follower's strategy.
We remark that our method is flexible and can incorporate even more designs to the follower's strategy.

We evaluate the performance of differentiable self-training on a set of weakly- and semi-supervised text classification and named entity recognition tasks. In some weakly-supervised learning tasks, our proposed method achieves competitive performance in comparison with fully-supervised models. For example, we obtain a 97.3\% vs. 96.2\% classification accuracy on Yelp, and we do not use any labeled training data from the Yelp dataset.

We highlight that our proposed differentiable self-training approach is an efficient substitution to existing self-training methods. Moreover, our method does not introduce any additional tuning parameter to the teacher-student framework. Additionally, {\ours} is flexible and can combine with various neural architectures.
We summarize our contributions as the following:
(1) We propose a differentiable self-training framework {\ours}, which employs a Stackelberg game formulation of the teacher-student approach.
(2) We employ differentiable pseudo-labels and differentiable sample weights as the follower's strategy. Our method alleviates the training instability issue.
(3) Extensive experiments on semi-supervised node classification, semi- and weakly-supervised text classification and named entity recognition tasks verify the efficacy of {\ours}.


%% file: 0-background.tex
\section{Background}




\noindent $\diamond$
\textbf{Self-training for semi-supervised learning. } Self-training is one of the earliest and simplest approaches to semi-supervised learning~\cite{rosenberg2005semi,lee2013pseudo}.
The method uses a teacher model to generate new labels, on which a student model is fitted.
Similar methods such as self-knowledge distillation~\cite{furlanello2018born} are proposed for supervised learning.
The major drawback of self-training is that it is vulnerable to label noise. A popular approach to tackle this is sample reweighting~\cite{freund1997decision, kumar2010self, malisiewicz2011ensemble}, where high-confidence samples~\cite{rosenberg2005semi, zhou2012self} are assigned larger weights. Data augmentation methods~\cite{berthelot2019mixmatch,chen2020mixtext} are also proposed to further enhance self-training.

\vspace{0.1in}
\noindent $\diamond$
\textbf{Self-training for weakly-supervised learning. }
Weak supervision sources, such as semantic rules and knowledge bases, facilitate generating large amounts of labeled data~\cite{goh2018using,hoffmann2011knowledge}.
The weak supervision sources have limited coverage, i.e., not all samples can be matched by the rules, such that a considerable amount of samples are unlabeled. Moreover, the generated weak labels usually contain excessive noise.
Recently, self-training techniques are adopted to weakly-supervised learning. In conjunction with pre-trained language models~\cite{devlin2018bert, liu2019roberta}, the technique achieves superior performance in various tasks~\cite{meng2018weakly, meng2020text,niu2020self, liang2020bond, yu2021fine}.

%% file: 0-model.tex
\section{Method}

For both semi-supervised and weakly-supervised learning problems, we have labeled samples $\cX_l=\{(x_i, y_i)\}_{i=1}^{N_l}$ and unlabeled samples $\cX_u=\{x_j\}_{j=1}^{N_u}$. Here $N_l$ is the number of labeled data, and $N_u$ is the number of unlabeled data.
Note that in weakly-supervised learning, we have unlabeled data because of the limited coverage of weak supervision sources.
The difference between semi- and weakly-supervised learning is that in the former case, the labels $\{y_i\}_{i=1}^{N_l}$ are assumed to be accurate, whereas in the latter case, the labels are noisy.
The goal is to learn a classifier $f: \cX \rightarrow \RR^C$, where $\cX=\cX_l \cup \cX_u$ denotes all the data samples, $\cY=\{1,\cdots, C\}$ is the label set, and $C$ is the number of classes.
The classifier $f$ outputs a point in the $C$-dimensional probability simplex, where each dimension denotes the probability that the input belongs to a specific class.

\subsection{Differentiable Self-Training for Semi-Supervised Learning}
Self-training can be interpreted as a teacher-student framework.
Within this framework, the teacher first generates pseudo-labels $\ty$ (see \eqref{eq:soft-label}) for the data samples. Then, the student updates itself by minimizing a loss function (see \eqref{eq:student-loss}), subject to the generated pseudo-labels.
Such two procedures are run iteratively.


We remark that self-training behaves poorly when encountering unreliable pseudo-labels, which will cause the student model to be updated towards the wrong direction.
To alleviate this issue, we find a good initialization $\theta\textsuperscript{init}$ for the models.
In semi-supervised learning, $\theta\textsuperscript{init}$ is found by fitting a model on the labeled data $\cX_l$.
Concretely, we solve
\begin{align} \label{eq:adaptation}
    \min_\theta ~\cL\textsubscript{sup}(\theta) = \frac{1}{N_l} \sum_{\cX_l} \ell\textsubscript{sup} \left( f(x_i, \theta), y_i \right).
\end{align}
Here $(x_i,y_i) \in \cX_l$, and $\ell\textsubscript{sup}(\cdot, \cdot)$ is the supervised loss, e.g., the cross-entropy loss.
\eqref{eq:adaptation} can be efficiently optimized using stochastic gradient-type algorithms, such as Adam~\cite{kingma2014adam}.

\begin{algorithm}[tb!]
\SetAlgoLined
\caption{Differentiable Self-Training.}
\label{alg:diff-self-train}
\KwIn{$\cX_l$: labeled dataset; $\cX_u$: unlabeled dataset; $\alpha$: parameter of exponential moving average; $\theta\textsuperscript{init}$: initialization; Optimizer: optimizer to update $\theta^S$.}
\textbf{Initialize:} $\theta^T_0=\theta^S_0=\theta\textsuperscript{init}$\;
\For{$t = 1, \cdots, T-1$}{
Sample a labeled minibatch $\cB_l=\{x_i\}_{i=1}^{|\cB_l|}$ from $\cX_l$\;
Sample an unlabeled minibatch $\cB_u=\{x_i\}_{i=1}^{|\cB_u|}$ from $\cX_u$\;
$\ty(\theta^T_t(\theta^S_t)) \leftarrow$ \eqref{eq:soft-label} on $\cB_u$\;
$\omega(\theta^T_t(\theta^S_t)) \leftarrow$ \eqref{eq:sample-weight} on $\cB_u$\;
$\cL(\theta^S_t) \leftarrow$ \eqref{eq:student-loss} on $\cB_u \cup \cB_l$\;
$g = \dd \cL(\theta^S_t) / \dd \theta^S_t \leftarrow$ \eqref{eq:stackelberg-gradient}\;
$\theta^S_{t+1} = \text{Optimizer}(\theta^S_t, g)$\;
$\theta^T_{t+1} = \alpha \theta^T_t + (1-\alpha) \theta^S_{t+1}$\;
}
\KwOut{Student model $\theta^S_{T}$ for prediction.}
\end{algorithm}

At time $t$, denote the student's parameters $\theta^S_t$, and the teacher's parameters $\theta^T_t(\theta^S_t)$.
We set both the student's and the teacher's initial parameters to $\theta\textsuperscript{init}$, i.e., $\theta_0^S = \theta_0^T(\theta_0^S) = \theta\textsuperscript{init}$.
Note that the teacher model depends on the student.
We adopt an exponential moving average~\cite{laine2016temporal, tarvainen2017mean} approach to model such a dependency:
\begin{align} \label{eq:mean-teacher}
    \theta^T_t(\theta^S_t) = \alpha \theta^T_{t-1} + (1-\alpha) \theta^S_t.
\end{align}

Recall that in our differentiable self-training framework, the student acknowledges the teacher's strategy. This meets the definition of a Stackelberg game~\cite{von2010market}, and we propose the following formulation:
\begin{align} \label{eq:diff-self-train}
    & \min_{\theta^S_t} \cL(\theta^S_t) = \cL\textsubscript{sup}(\theta_t^S) \\
    & \quad\quad\quad + \frac{1}{N_u} \sum_{x_i \in \cX_u} \ell_S \left( x_i, F(\theta^T_t(\theta^S_t)), \theta^S_t \right), \notag \\
    & \text{s.t. } F\left( \theta^T_t(\theta^S_t) \right) = \left[ \ty(\theta^T_t(\theta^S_t)), \omega(\theta^T_t(\theta^S_t)) \right]. \notag
\end{align}
Here recall that $\cX_u$ is the unlabeled data samples, and $N_u$ is the size of $\cX_u$. In \eqref{eq:diff-self-train}, $F(\theta^T_t(\theta^S_t))$ is the teacher's strategy, which contains differentiable pseudo-labels (i.e., $\ty(\theta^T_t)$ in \eqref{eq:soft-label}) and differentiable sample weights (i.e., $\omega(\theta^T_t)$ in \eqref{eq:sample-weight}). The loss function $\ell_S$ is defined in \eqref{eq:student-loss}. Note that we still include the supervised loss $\cL\textsubscript{sup}$ in \eqref{eq:adaptation} in the objective function $\cL$.
Following conventions, in \eqref{eq:diff-self-train}, the minimization problem solves for the leader, and we call $F(\theta^T_t)$ the follower's strategy. Note that the Stackelberg game formulation \eqref{eq:diff-self-train} has also been adopted in adversarial training \citep{zuo2021adversarial}.

The Stackelberg game formulation is fundamentally different from conventional self-training approaches, where the teacher $\theta^T$ is not treated as a function of the student $\theta^S$.
In our differentiable self-training framework, the leader takes the follower's strategy into account by considering $F(\theta^T_t(\theta^S_t))$.
In this way, self-training can be viewed solely in terms of the leader's parameters $\theta^S_t$.

Consequently, the leader problem can be efficiently solved using stochastic gradient-type algorithms, where the gradient is
\begin{align} \label{eq:stackelberg-gradient}
    & \frac{\dd \cL(\theta^S_t)}{\dd \theta^S_t}
    = \frac{1}{N_l} \sum_{(x_i,y_i) \in \cX_l} \frac{\dd \ell\textsubscript{sup}(\theta_t^S)}{\dd \theta_t^S} \\
    & + \frac{1}{N_u}\sum_{x_i \in \cX_u} \frac{\dd \ell_S \left( x_i, F(\theta^T_t(\theta^S_t)), \theta^S_t \right)}{\dd \theta^S_t} \notag \\
    & = \underbrace{\frac{1}{N_l} \sum_{\cX_l} \frac{\dd \ell\textsubscript{sup}(\theta_t^S)}{\dd \theta_t^S} + \frac{1}{N_u}\sum_{\cX_u}\frac{\partial \ell_S \left(x_i, F, \theta^S_t \right)}{\partial \theta^S_t}}_{\text{leader}} \notag \\[5pt]
    & + \underbrace{\frac{1}{N_u}\sum_{x_i\in \cX_u} \frac{\partial \ell_S \left(x_i, F(\theta^T_t(\theta^S_t)), \theta^S_t \right)}{\partial \theta^T_t(\theta^S_t)} \frac{\dd \theta^T_t(\theta^S_t)}{\dd \theta^S_t}}_{\text{leader-follower interaction}}. \notag
\end{align}
In \eqref{eq:stackelberg-gradient}\footnote{The ``leader'' term is written as $\partial \ell_S(x_i, F, \theta^S_t) / \partial \theta^S_t$ instead of $\partial \ell_S(x_i, F(\theta^T_t(\theta^S_t)), \theta^S_t) / \partial \theta^S_t$ because the partial derivative is only taken with respect to the third argument in $\ell_S(x_i, F, \theta^S_t)$. We drop the $\theta^T_t(\theta^S_t)$ term in $F(\theta^T_t(\theta^S_t))$ to avoid causing confusion.}, we have $\dd \theta^T_t(\theta^S_t) / \dd \theta^S_t = 1-\alpha$ because of \eqref{eq:mean-teacher}.
Note that a conventional self-training method only considers the ``leader'' term, and ignores ``leader-follower interaction''.
This causes training instabilities, which we demonstrate empirically in Fig.~\ref{fig:two-moon-learning} and Fig.~\ref{fig:two-moon}.

The proposed differential self-training algorithm is summarized in Algorithm~\ref{alg:diff-self-train}.
In the next two sections, we spell out the two components of the follower's strategy, namely differentiable pseudo-labels and differentiable sample weights.

We remark that Algorithm~\ref{alg:diff-self-train} adopts a Stackelberg game formulation of self-training. That is, the loss terms in \eqref{eq:diff-self-train} (soft-labels and sample weights) are well-established techniques, and the proposed method is a novel optimization algorithm.


\subsection{Differentiable Pseudo-Labels}
\label{sec:diff-label}

In a self-training framework, the teacher model labels the unlabeled data. Concretely, at time $t$, for each sample $x \in \cX_u$ in the unlabeled dataset, a hard pseudo-label~\citep{lee2013pseudo} is defined as
\begin{align} \label{eq:hard-label}
    \tilde{y}_\textsubscript{hard} = \underset{j \in \cY}{\mathrm{argmax}} \left[ f(x, \theta^T_t) \right]_j.
\end{align}
Here $f(x,\theta^T_t) \in \RR^C$ is in the probability simplex, and $[f(x,\theta^T_t)]_j$ denotes its $j$-th entry.

There are two problems with the hard pseudo-labels. First, differentiable self-training requires every component of the follower's strategy \eqref{eq:diff-self-train} to be differentiable with respect to the leader's parameters. However, \eqref{eq:hard-label} introduces a non-differentiable $\mathrm{argmax}$ operation.
Second, the hard pseudo-labels exacerbates error accumulation. This is because $\tilde{y}_\textsubscript{hard}$ only contains information about the most likely class, such that statistics regarding the prediction confidence $f(x, \theta^T_t)$ is lost. For example, suppose in a two-class classification problem, we obtain $f(x, \theta^T_t)=[0.51, 0.49]$ for some $x$. This prediction result indicates that the model is uncertain to which class $x$ belongs. However, under the hard pseudo-label $\tilde{y}_\textsubscript{hard}=0$, the student model becomes unaware of such uncertainty.

To resolve the above two issues, we propose to employ soft pseudo-labels~\cite{xie2016unsupervised, xie2020uda, meng2020text}. Concretely, for a data sample $x \in \cB$ in a batch $\cB$, the $j$-th entry of its soft pseudo-label $\ty(\theta^T_t) \in \RR^C$ is defined as
\begin{align} \label{eq:soft-label}
    \left[ \ty(\theta^T_t) \right]_j = \frac{\left[ f(x, \theta^T_t(\theta^S_t)) \right]^{1/\tau} / f_j}{\sum_{j' \in \cY} \left[ f(x, \theta^T_t(\theta^S_t)) \right]^{1/\tau} / f_{j'}},
\end{align}
where $f_j = \sum_{x' \in \cB} [f(x', \theta^T_t(\theta^S_t))]^{1/\tau}$, and $\tau$ is a temperature parameter that controls the ``softness'' of the soft pseudo-label. Note that when the temperature is low, i.e., $\tau \rightarrow 0$, the soft pseudo-label becomes sharper and eventually converges to the hard pseudo-label \eqref{eq:hard-label}.

In \eqref{eq:soft-label}, the soft pseudo-label $\ty(\theta^T_t)$ is a function of the teacher's parameters $\theta^T_t$, which in turn is a function of the student's parameters $\theta^S_t$ \eqref{eq:mean-teacher}. Therefore, $\ty$ is differentiable with respect to $\theta^S_t$, and fits in the differentiable self-training framework.
The gradient of $\tilde{y}$ with respect to $\theta^S$ can be efficiently computed by a single back-propagation using deep learning libraries.

Notice that \eqref{eq:soft-label} emphasizes the tendency of $x$ belonging to a specific class, instead of to which class $x$ belongs. Therefore, even when $\tilde{y}_\textsubscript{hard}$ is wrong, the soft version of it is still informative.

\subsection{Differentiable Sample Weights}
\label{sec:diff-weight}

Sample reweighting is an effective tool to tackle erroneous labels~\cite{freund1997decision, kumar2010self, malisiewicz2011ensemble, liang2020bond}. Specifically, pseudo-labels that have dominating entries are more likely to be accurate than those with uniformly distributed entries.
For example, a sample labeled $[0.9,0.1]$ is more likely to be classified correctly than a sample labeled $[0.6,0.4]$. With this intuition, for a sample and its soft pseudo-label $\ty(\theta^T_t)$, we define its sample weight $\omega$ as
\begin{align} \label{eq:sample-weight}
    \omega(\theta^T_t) = 1- \frac{H \left(\ty(\theta^T_t) \right)}{\log (C)},  
\end{align}
where $H(\ty(\theta^T_t)) = -\sum_{j=1}^C \ty_j \log (\ty_j)$ is the entropy of $\ty(\theta^T_t)$ that satisfies $0 \leq H(\ty(\theta^T_t)) \leq \log(C)$.
Note that if the pseudo-label is uniformly distributed, then the corresponding sample weight is low, and vice versa.
Similar to the pseudo-label $\ty(\theta^T_s)$, the sample weight $\omega(\theta^T_t)$ is a function of the teacher's parameters $\theta^T_t$, and further a function of the student's parameters $\theta^S_t$.

With the differentiable pseudo-labels and the differentiable sample weights, we define the student's loss function as
\begin{multline} \label{eq:student-loss}
    \ell_S \left( x_i, F(\theta^T_t(\theta^S_t)), \theta^S_t \right)
    = \omega \left( \theta^T_t(\theta^S_t) \right) \\
    \mathrm{KL}\left( \ty(\theta^T_t(\theta^S_t)) \| f(x_i, \theta^S_t) \right),
\end{multline}
where $\mathrm{KL}(p||q)=\sum_k p_k \log(p_k/q_k)$ is the Kullback–Leibler (KL) divergence.

\subsection{Weakly-Supervised Learning}
Recall that in weakly-supervised learning, we have both labeled data $\cX_l$ and unlabeled data $\cX_u$. 
Note that weak supervision sources often yield noisy labels. This is because data are annotated automatically by, for example, linguistic rules, which have limited accuracy. As such, the supervised loss $\cL\textsubscript{sup}$ in \eqref{eq:diff-self-train} only exacerbates the label noise issue.

We address this problem by discarding the noisy weak labels in $\cX_l$ after obtaining the initialization $\theta\textsuperscript{init}$ \eqref{eq:adaptation}.
Accordingly, we adopt the following formulation for weakly-supervised learning:
\begin{align} \label{eq:weak-formulation}
    & \min_{\theta^S_t} \cL(\theta^S_t) = \frac{1}{N} \sum_{x_i \in \cX_u \cup \cX_l} \ell_S \left( x_i, F, \theta^S_t \right), \\
    & \text{s.t. } F\left( \theta^T_t(\theta^S_t) \right) = \left[ \ty(\theta^T_t(\theta^S_t)), \omega(\theta^T_t(\theta^S_t)) \right]. \notag
\end{align}
Here $N=N_l+N_u$ is the total number of training samples. Note that in comparison with \eqref{eq:diff-self-train}, we drop the supervised loss $\cL\textsubscript{sup}$.
Moreover, the teacher model now generates soft pseudo-labels for all the data, instead of only the data in $\cX_u$.

%% file: 0-experiments.tex
\section{Experiments}
We conduct two sets of experiments: weakly- and semi-supervised text classification. We also examine semi-supervised node classification on graphs (Appendix~\ref{app:graph}).
All the results have passed a paired t-test with $p<0.05$.
When using pre-trained language models, we employ a RoBERTa-base~\cite{liu2019roberta} model obtained from the HuggingFace~\cite{wolf2019huggingface} codebase.
We implement all the methods using PyTorch~\citep{paszke2019pytorch}, and experiments are run on NVIDIA 2080Ti GPUs.
All the training details are deferred to the appendix.

\begin{figure}[tb!]
    \centering
    \includegraphics[width=1.0\linewidth]{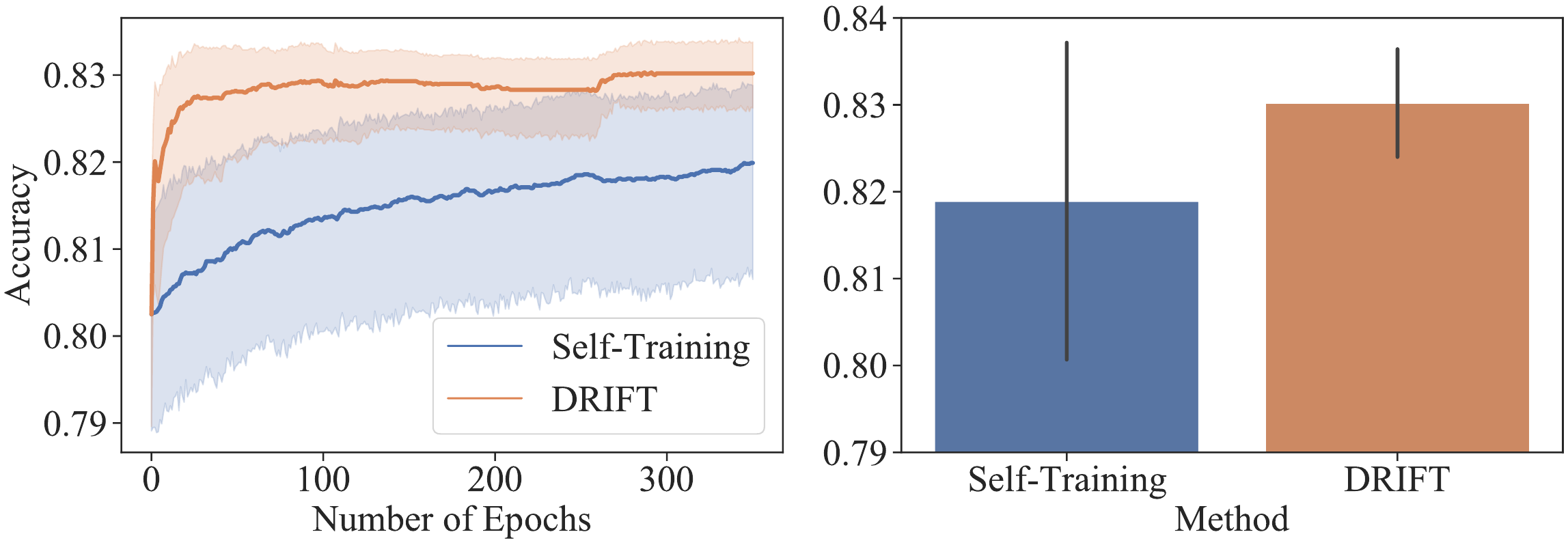}
    \vskip -0.05in
    \caption{$\textit{Left}$: Learning curve. $\textit{Right}$: The last epoch accuracy. The vertical lines signify run-to-run variance.}
    \label{fig:two-moon-learning}
\end{figure}
\begin{figure}[tb!]
    \centering
    \includegraphics[width=1.0\linewidth]{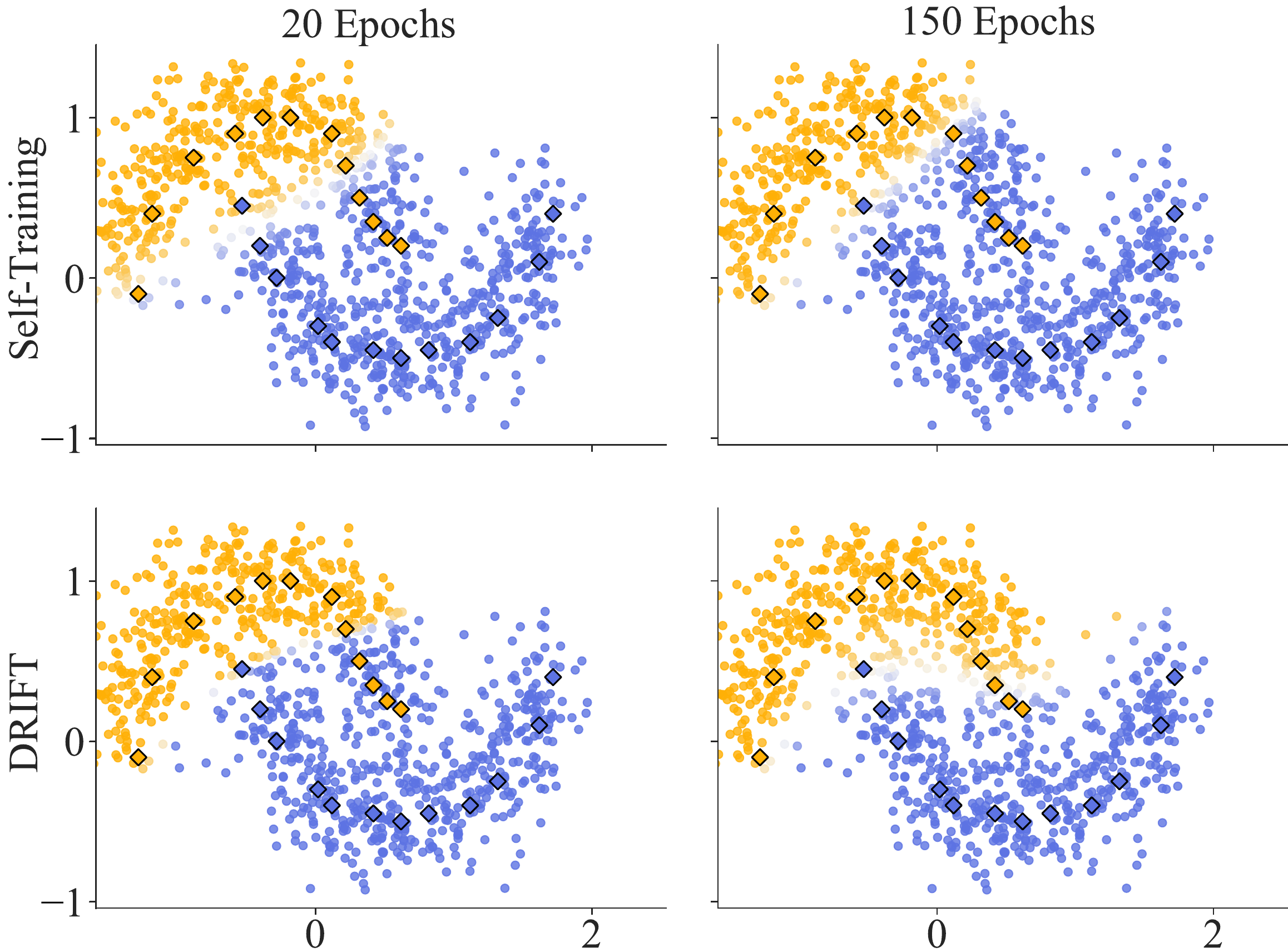}
    \vskip -0.05in
    \caption{Demonstration of Self-training stuck at bad optima. The solid diamonds are labeled samples.}
    \label{fig:two-moon}
\end{figure}

\newcolumntype{C}{@{\hskip3pt}c@{\hskip3pt}}
\newcolumntype{L}{l@{\hskip2pt}}
\begin{table*}[tb!]
\centering \small
\resizebox{2.1\columnwidth}{!}{
\begin{tabular}{L|C|C|C|C|C|C|C|C}
\toprule
\textbf{Dataset} & \ \bf AGNews \ & \ \bf IMDB \ & \ \bf Yelp \ & \ \bf MIT-R \ &  \bf CoNLL-03 \ & \ \bf Webpage \ & \ \bf BC5CDR \ & \ \bf Wikigold \ \\ \midrule
RoBERTa-Full & 91.41 & 94.26 & 97.27 & 88.51 & 90.11~(89.14/91.10) & 72.39~(66.29/79.73) & 85.15~(83.74/86.61) & 86.43~(85.33/87.56) \\ \midrule
RoBERTa-Weak & 82.25 & 72.60 & 79.91 & 70.95 & 75.61~(\textbf{83.76}/68.90) & 59.11~(60.14/58.11) & 78.51~(74.96/82.42) & 51.55~(49.17/54.50) \\
WeSTClass & 82.78 & 77.40 & 76.86 & -\!-\!- & -\!-\!- & -\!-\!- & -\!-\!- & -\!-\!- \\
Self-training & 86.07 & 85.72 & 89.95 & 73.59 & 77.28~(83.42/71.98) & 56.90~(54.32/59.74) & 79.92~(74.73/\textbf{85.90}) & 56.90~(54.32/59.74) \\
UAST & 86.28 & 84.56 & 90.53 & 74.41 & 77.92~(83.30/73.20) & 58.18~(56.33/60.14) & 81.50~(80.09/82.98) & 57.79~(52.64/64.05)\\
BOND & 86.19 & 88.36 & 93.18 & 75.90 & 81.48~(82.05/80.92) & 65.74~(\textbf{67.37}/64.19) & 81.53~(79.54/83.63) & 60.07~(53.44/\textbf{68.58}) \\
\midrule
\ours & \textbf{87.80} & \textbf{91.56} & \textbf{96.24} & \textbf{77.15} & \textbf{81.74}~(81.45/\textbf{82.02}) & \textbf{66.04}~(65.23/\textbf{66.87}) & \textbf{82.62}~(\textbf{82.57}/82.68) & \textbf{60.66}~(\textbf{57.50}/64.21) \\
\bottomrule
\end{tabular}
}
\vspace{-0.05in}
\caption{Accuracy (in \%) of weakly-supervised text classification on various datasets. We report the mean over three runs. {\ours} is initialized from RoBERTa-Weak. For text classification tasks, we report the accuracy; and for NER tasks, we report $F_1$ (precision/recall). The best results are shown in bold, except \textit{RoBERTa-Full}, which is a fully-supervised model and is included here as a reference.}
\label{tb:weak-results}
\end{table*}

\subsection{Warmup: TwoMoon Experiments}
To understand the efficacy of {\ours}, we conduct a semi-supervised classification experiment on a classic synthetic dataset ``TwoMoon''. The dataset contains two classes, and for each class we generate $12$ labeled samples and $500$ unlabeled ones. 

We compare {\ours} with conventional Self-training. The only difference between the two methods is that {\ours} adopts the differentiable strategies, while Self-training does not.
In both methods, the teacher/student model is a two-layer feed-forward neural network, with hidden dimension $50$ and $\mathrm{tanh}$ (hyperbolic tangent) as the non-linearity.
We first train the models for $50$ epochs using the labeled samples. We then conduct self-training with learning rate 0.01 and Adam \citep{kingma2014adam} as the optimizer.
We adopt an exponential moving average approach \eqref{eq:mean-teacher} with $\alpha=0.5$, and we set the temperature parameter $\tau=0.5$ for the soft pseudo-labels \eqref{eq:soft-label}. 

We conduct $10$ trails, and Fig.~\ref{fig:two-moon-learning} shows the accuracy and the variance during training. We can see that Self-training yields a much larger variance, indicating an unstable training process.
Note that the performance gain of {\ours} to Self-training has passed a paired-student t-test with p-value $<0.05$.

Moreover, by examining the experimental results, we find that Self-training at times gets stuck at subpotimal solutions.
As an example, in Fig.~\ref{fig:two-moon}, notice that the two methods behave equally well at epoch 20. However, Self-training gets stuck and does not improve at epoch 150. This is because the teacher generates hazardous labels that avert the student from improving.
Meanwhile, by incorporating differentiable strategies, the performance of {\ours} improves at epoch 150 from epoch 20.

\subsection{Weakly-Supervised Text Classification}
We fine-tune a pre-trained RoBERTa model for weakly-supervised learning. In addition, we demonstrate that our method works well when trained-from-scratch and when using different backbones than the Transformer~\cite{vaswani2017attention}. See Section~\ref{sec:semi} and Table~\ref{tb:encoder} for details.

\vspace{0.1in}
\noindent
\textbf{Settings. }
We use the following datasets: Topic Classification on AGNews~\cite{zhang2015character}; Sentiment Analysis on IMDB~\cite{IMDB} and Yelp~\cite{meng2018weakly}; Slot Filling on MIT-R~\cite{liu2013query}; and Named Entity Recognition (NER) on CoNLL-03~\cite{sang2003introduction}, Webpage~\cite{ratinov2009design}, Wikigold~\cite{balasuriya2009named}, and BC5CDR~\cite{li2016biocreative}.
The dataset statistics are summarized in Table~\ref{tb:text-dataset}.
For each dataset, we generate weak labels using some pre-defined rules, after which the same data and generated weak labels are used by all the methods.
More details about the weak supervision sources are in Appendix~\ref{app:dataset}.

We adopt several baselines:
\begin{itemize}
    \item \textit{RoBERTa}~\cite{liu2019roberta} uses the RoBERTa-base model with task-specific classification heads.
    \item \textit{Self-training}~\cite{lee2013pseudo,rosenberg2005semi} uses the conventional teacher-student framework, where a teacher generates pseudo-labels, and a student makes predictions.
    \item \textit{WeSTClass}~\cite{meng2018weakly} leverages generated pseudo-documents and uses self-training to bootstrap over all the samples. 
    \item \textit{BOND}~\cite{liang2020bond} uses a teacher-student framework for self-training. The teacher model is periodically updated to generate pseudo-labels when training the student.
    \item \textit{UAST}~\cite{mukherjee2020uncertainty} estimates uncertainties of unlabeled data via MC-dropout~\cite{gal2016dropout} during self-training, and then selects samples with low uncertainties. It is the state-of-the-art self-training method for text data with few labels.
\end{itemize}

Recall that for weakly-supervised learning, we first fine-tune a RoBERTa model using the weakly-labeled data, and then we discard the weak labels and continue with self-training. This is an effective strategy to reduce overfitting on label noise \citep{yu2021fine}. We follow this procedure for both {\ours} and all the baseline methods.

\vspace{0.1in}
\noindent \textbf{Results. }
Experimental results are summarized in Table~\ref{tb:weak-results}. We can see that {\ours} achieves the best performance in all the tasks. Notice that the baselines that adopt self-training, e.g., WestClass, Self-training, UAST, and BOND, outperform the vanilla RoBERTa-Weak method. This is because in weakly-supervised learning, a noticeable amount of labels are inaccurate. Therefore, without noise suppressing approaches such as self-training, models cannot behave well.
However, without taking the teacher's strategy into account, these methods still suffer from training instabilities, such that they are not as effective as {\ours}.

We highlight that on some datasets, performance of our method is close to the fully-supervised model RoBERTa-Full, even though we do not use any clean labels. For example, {\ours} achieves 91.6\% vs. 94.3\% performance on IMDB, 96.2\% vs. 97.3\% on Yelp, and 82.6 vs. 85.1 on BC5CDR.

\subsection{Semi-Supervised Text Classification}
\label{sec:semi}

\begin{table*}[htb!]
\centering \small
\begin{tabular}{L|cccc|cccc|cccc}
\toprule
\textbf{Dataset} & \multicolumn{4}{c|}{\textbf{AGNews}} & \multicolumn{4}{c|}{\bf IMDB} &   \multicolumn{4}{c}{\bf Amazon} \\ \cline{1-1}
\textbf{Labels/class} & 30 & 50 & 200 & 1000 & 30 & 50 & 200 & 1000 & 30 & 50 & 200 & 1000 \\
\midrule
RoBERTa-Semi & 83.98 & 87.44 & 88.01 & 90.91 & 86.64 & 88.37 & 89.25 & 90.54 & 88.21 & 89.66 & 92.31 & 93.65 \\
VAMPIRE & -\!-\!- & -\!-\!- & 83.90 & 85.80 & -\!-\!- & -\!-\!- & 82.20 & 85.40 & -\!-\!- & -\!-\!-&-\!-\!- & -\!-\!- \\
UDA & 85.92 & 88.09 & 88.33 & 91.22 & 89.30 & 89.42 & 89.72 & 90.87 & -\!-\!- & -\!-\!- & -\!-\!- & -\!-\!- \\
MixText & 88.50 & 88.85 & 89.20 & 91.55 & 84.34 & 88.72 & 89.45 & 91.20 & -\!-\!- & -\!-\!- & -\!-\!- & -\!-\!-\\
Self-training & 84.62 & 88.04 & 88.67 & 91.47 & 88.13 & 88.80 & 89.84 & 91.04 & 89.92 & 90.55 & 92.55 & 93.83 \\
UAST & 87.74 & 88.65 & 89.21 & 91.81 & 89.21 & 89.56 & 90.11 & 91.48 & 91.27 & 91.50 & 92.68 & 93.97 \\ \midrule
{\ours} & \textbf{89.46} & \textbf{89.67} & \textbf{90.17} & \textbf{92.47} & \textbf{89.77} & \textbf{90.03} & \textbf{90.83} & \textbf{92.39} & \textbf{91.82} & \textbf{92.67} & \textbf{93.16} & \textbf{94.28} \\
\bottomrule
\end{tabular}
\vspace{-0.05in}
\caption{Accuracy (in \%) of semi-supervised text classification on various datasets. {\ours} is initialized from RoBERTa-Semi. The best results are shown in bold.}
\label{tb:semi-results}
\end{table*}

\begin{table*}[htb!]
\centering \small
\begin{tabular}{l|ccccc|ccccc}
\toprule
\textbf{Dataset} & \multicolumn{5}{c|}{\textbf{AGNews}} & \multicolumn{5}{c}{\textbf{IMDB}} \\ \cline{1-1}
\textbf{Labels per class} & weak & 30 & 50 & 200 & 1000 & weak & 30 & 50 & 200 & 1000 \\ \midrule
TextCNN & 79.45 & 78.81 & 79.98 & 85.46 & 86.78 & 82.44 & 63.32 & 66.61 & 73.22 & 78.29 \\
Self-training & 81.69 & 81.98 & 82.67 & 86.26 & 88.15 & 84.76 & 64.68 & 65.26 & 73.60 & 79.04 \\
UAST & 81.48 & 82.05 & 83.34 & 86.67 & 87.90 & 83.97 & 64.23 & 68.70 & 73.95 & 79.13 \\ \midrule
\ours & \textbf{82.55} & \textbf{83.34} & \textbf{85.01} & \textbf{87.38} & \textbf{88.66} & \textbf{86.44} & \textbf{65.65} & \textbf{69.86} & \textbf{74.61} & \textbf{79.38} \\
\bottomrule
\end{tabular}
\vspace{-0.05in}
\caption{Results of {\ours} and self-training baselines on AGNews and IMDB. We use TextCNN as the backbone and train the models from scratch. {\ours} is initialized from TextCNN. The best results are shown in bold. ``Weak'' means the weak-supervision setting.}
\label{tb:encoder}
\end{table*}

\noindent \textbf{Datasets. }
We adopt AGNews, IMDB, and Amazon~\cite{mcauley2013hidden} (see Table~\ref{tb:text-dataset}) in this set of experiments. For each dataset, we randomly sample $N \in \{30,50,200,1000\}$ data points from each class and annotate them with clean labels, while the other data are treated as unlabeled. Note that for all the splits of a particular dataset, we use the same development and test sets.

\vspace{0.1in}
\noindent \textbf{Settings. }
Our differentiable self-training framework works well in both fine-tuning and training-from-scratch regimes. Moreover, our approach is flexible to accommodate different neural architectures. We conduct two sets of experiments.
In the first set, we fine-tune a pre-trained RoBERTa model, which uses the Transformer~\cite{vaswani2017attention} as its backbone.
In the second set of experiments, we train a TextCNN~\citep{kim2014convolutional} model from scratch, which employs a convolutional neural network as the foundation.

\vspace{0.1in}
\noindent \textbf{Baselines. }
Besides RoBERTa, Self-training, and UAST, which are used in weakly-supervised classification tasks, we adopt several new methods as baseline approaches.

\begin{itemize}
    \item \textit{VAMPIRE}~\cite{gururangan2019variational} pre-trains a unigram document model on unlabeled data using a variational auto-encoder, and then uses its internal states as features for downstream applications.
    \item \textit{UDA}~\cite{xie2020uda} uses back translation and word replacement to augment unlabeled data, and forces the model to make consistent predictions on the augmented data to improve model performance.
    \item \textit{MixText}~\cite{chen2020mixtext} augments the training data by interpolation in the hidden space, and it exploits entropy and consistency regularization to further utilize unlabeled data during training.
\end{itemize}

\vspace{0.1in}
\noindent \textbf{Results. } Experimental results are summarized in Table~\ref{tb:semi-results}.
We can see that {\ours} achieves the best performance across the three datasets under different setups.
Notice that the performance of VAMPIRE is not satisfactory. This is because it does not use pre-trained models, unlike the other baselines. Pre-trained language models contain rich semantic knowledge, which can be effectively transferred to the target task and boost model performance.
All the baselines do not explicitly consider the teacher's strategy, and thus, they suffer from training instabilities.

We remark that UDA, UAST and MixText leverage external sources or data augmentation methods to make full use of the unlabeled data. These methods can potentially combine with {\ours}, which is of separate interests.

\vspace{0.1in}
\noindent \textbf{Fine-tuning vs. Training-from-scratch. }
Table~\ref{tb:encoder} shows the results of training a TextCNN model from scratch.
We can see that the model trained from scratch performs worse than fine-tuning a pre-trained model (Table~\ref{tb:semi-results}). This is because TextCNN has significantly less parameters than RoBERTa, and is not pre-trained on massive text corpora. Therefore, we cannot take advantage of the semantic information from pre-trained models.

Nevertheless, under both weakly-supervised and semi-supervised learning settings, {\ours} consistently outperforms the baseline methods.
This indicates that our method is architecture independent, and does not rely on transferring existing semantic information. As such, differentiable self-training serves as an effective plug-in module for existing models. We remark that {\ours} does not introduce any additional tuning parameter in comparison with conventional self-training.

\subsection{Ablation Study}

\noindent $\diamond$
\textbf{Components of {\ours}. }
We inspect different components of {\ours}, including the differentiable pseudo labels (DrPL), differentiable sample weights (DrW), and the sample reweighting strategy (SR)\footnote{For models without DrPL, we do not differentiate the pseudo-labels. For models without DrW, we still use \eqref{eq:sample-weight} to perform sample rewriting, but we do not differentiate the weights, i.e., w/o DrW equals to w/ SR. For models without SR, we do not use sample reweighing.}.
Experimental results are summarized in Table~\ref{tb:ablation_study}.
We observe that both differentiable pseudo-labels and differentiable sample weights contribute to model performance, as removing any of them hurts the classification accuracy.
Also, {\ours} excels when the labels are noisy. We can see that our method brings 2.41\% performance gain on average under the weakly-supervised setting, while it only promotes 1.08\% average gain under the semi-supervised setting. Such results indicate that differentiable pseudo-labels and sample weight are effective in suppressing label noise.

\begin{table}[tb!]
\centering \small
\begin{tabular}{@{\hskip5pt}l@{\hskip5pt}|CCC|CCC}
\toprule
\textbf{Method} & \multicolumn{3}{c}{\textbf{AGNews}} &  \multicolumn{3}{c}{\textbf{IMDB}} \\ \midrule
\textbf{\#labels} & 30 & 1000 & weak & 30 & 1000 & weak \\ \midrule
{\ours} & \textbf{89.46} & \textbf{92.47} & \textbf{87.80} & \textbf{89.77} & \textbf{92.39} & \textbf{91.56}  \\
~~w/o DrPL & 87.63 & 92.11 & 86.13 & 89.05  & 91.75 & 86.78  \\
~~w/o DrW & 88.51  & 91.22 & 86.62 & 88.47 & 91.49  & 90.24  \\
~~w/o SR & 88.84  & 91.20 & 86.19 & 88.15 & 91.95 & 87.76 \\
\bottomrule
\end{tabular}
\vspace{-0.05in}
\caption{Effects of different components of \ours. Here ``weak'' means the weak-supervision setting.}
\label{tb:ablation_study}
\end{table}

\begin{figure}[tb!]
    \centering
    \begin{subfigure}[t]{0.49\linewidth}
         \centering
         \includegraphics[width=\linewidth]{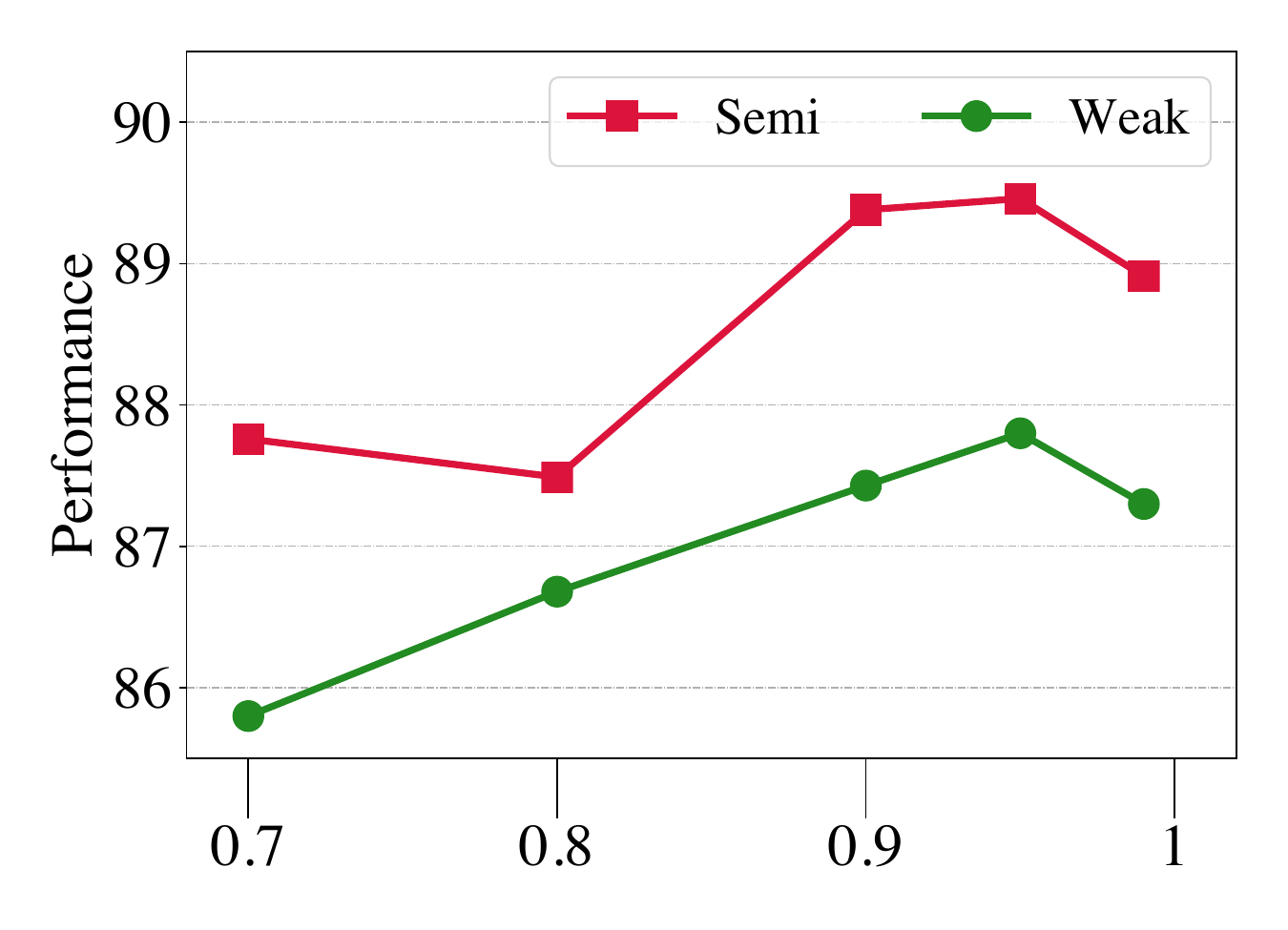}
         \vskip -0.05in
         \caption{Effect of $\alpha$ on AGNews.}
     \end{subfigure} \hfill
     \begin{subfigure}[t]{0.49\linewidth}
         \centering
         \includegraphics[width=\linewidth]{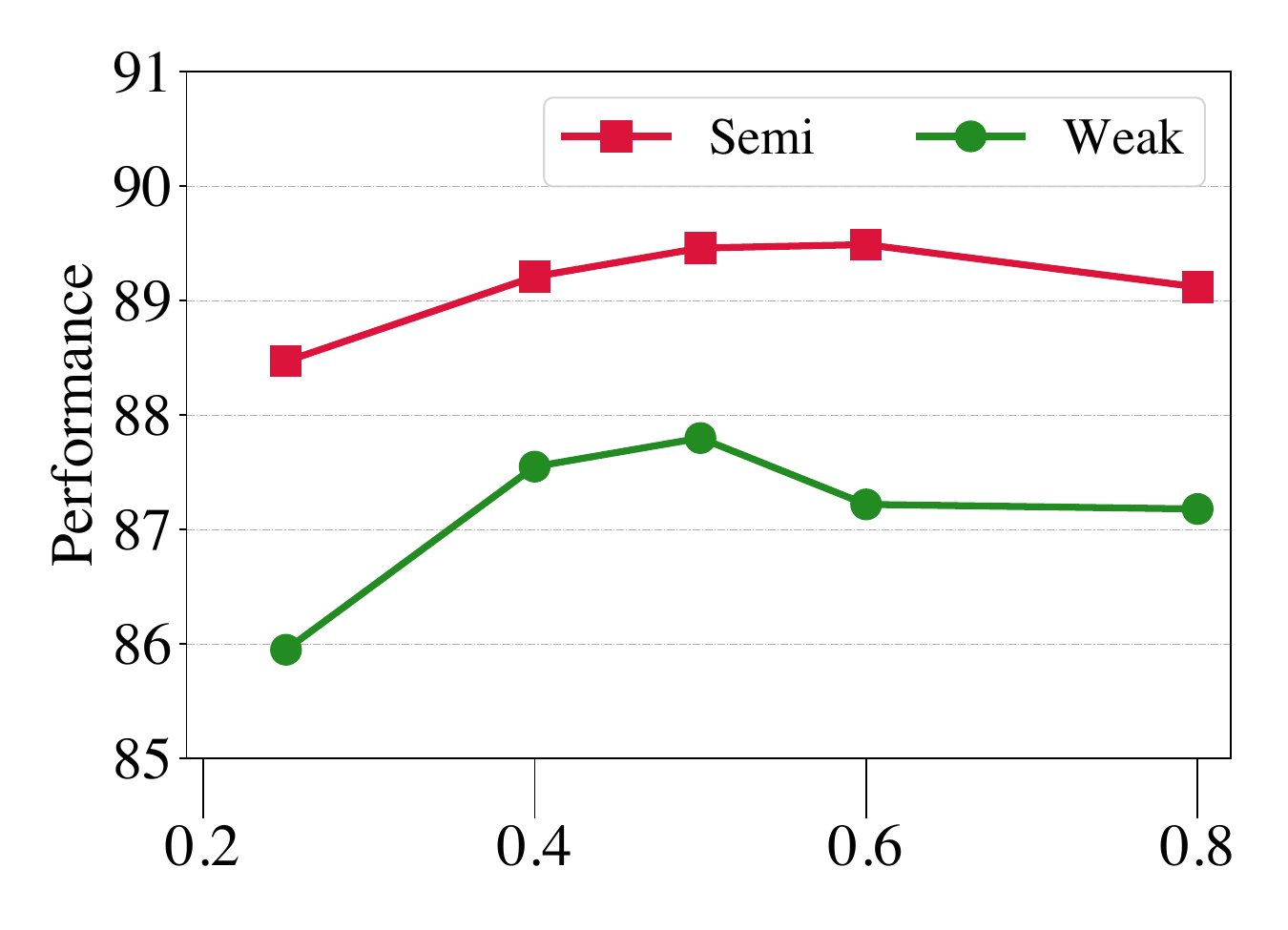}
         \vskip -0.05in
         \caption{Effect of $\tau$ on AGNews.}
     \end{subfigure}
     \begin{subfigure}[t]{0.49\linewidth}
         \centering
         \includegraphics[width=\linewidth]{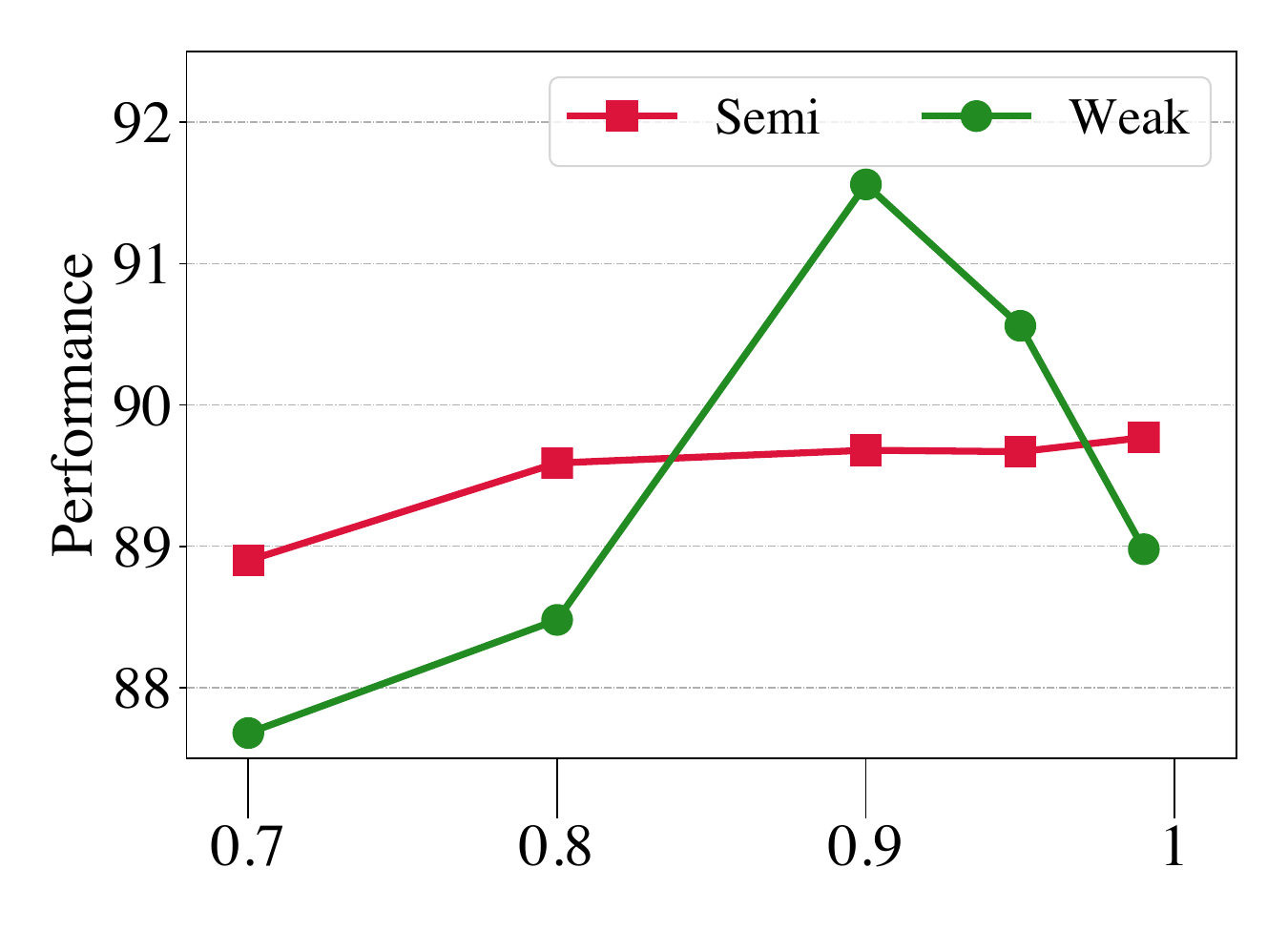}
         \vskip -0.05in
         \caption{Effect of $\alpha$ on IMDB.}
     \end{subfigure} \hfill
     \begin{subfigure}[t]{0.49\linewidth}
         \centering
         \includegraphics[width=\linewidth]{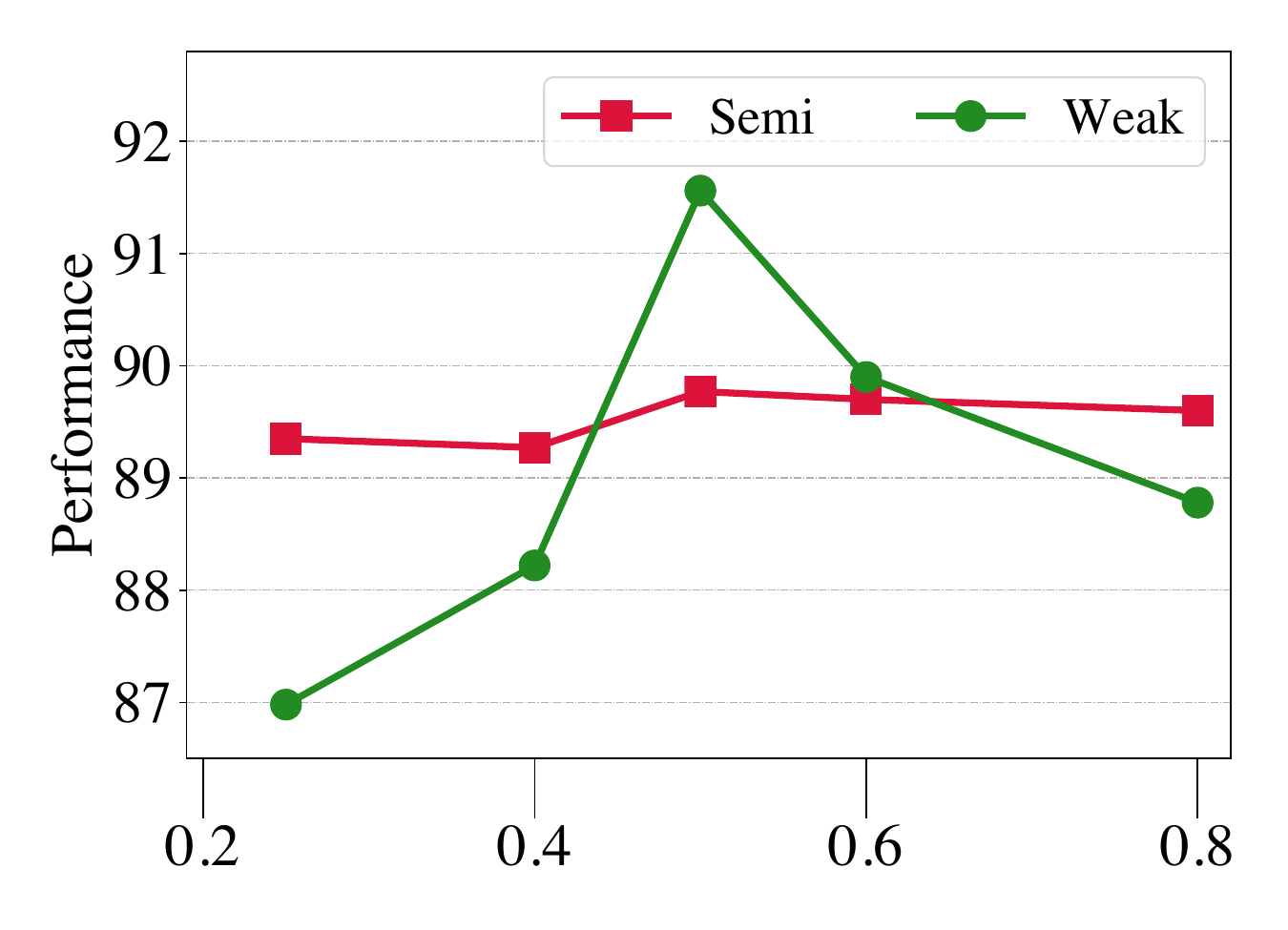}
         \vskip -0.05in
         \caption{Effect of $\tau$ on IMDB.}
     \end{subfigure}
     \vskip -0.05in
     \caption{Parameter study. Here ``semi'' means the semi-supervision setting with 30 labels per class, and ``weak'' means the weak-supervision setting.}
    \label{fig:para-study}
\end{figure}

\vspace{0.1in}
\noindent $\diamond$
\textbf{Sensitivity to hyper-parameters. }
We study models' sensitivity to the exponential moving average rate $\alpha$ and the soft pseudo-label's temperature $\tau$. Figure~\ref{fig:para-study} shows the results.
We can see that model performance peaks when $\alpha$ is around $0.9$. The teacher model updates too aggressively with a smaller $\alpha$ (e.g., $\alpha=0.7$), and too conservatively with a larger alpha (e.g., $\alpha=0.99$). In the first case, the generated pseudo-labels are not reliable; and in the second case, model improves too slow.
Also notice that the semi-supervised model is not sensitive to the temperature parameter. The weakly-supervised model achieves the best performance when $\tau=0.5$. Note that a smaller $\tau$ essentially generates hard pseudo-labels, which drastically hurts model performance.

\subsection{Case Study}

\begin{figure}[tb!]
    \centering
    \includegraphics[width=\linewidth]{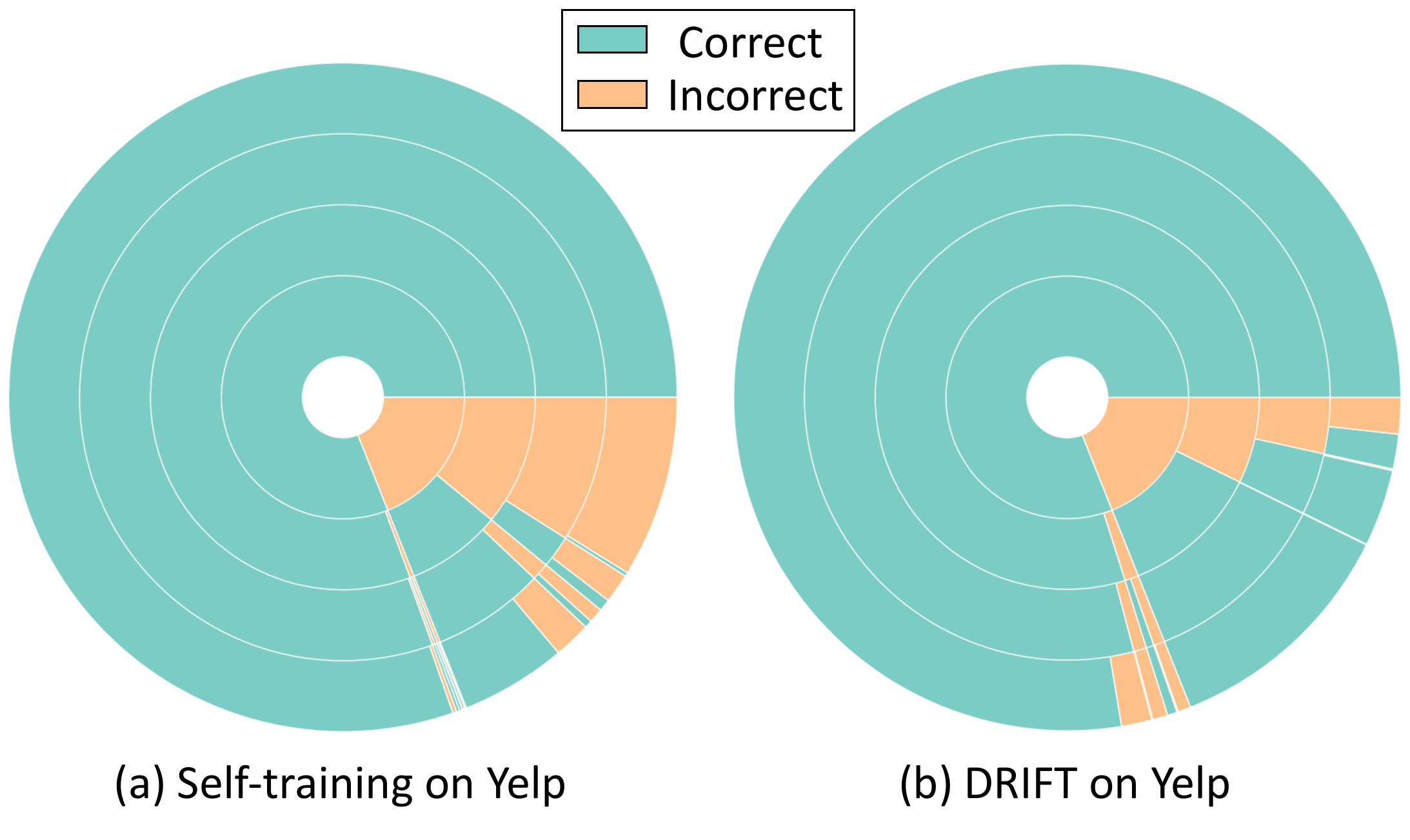}
    \vskip -0.05in
    \caption{Sample predictions under weak supervision. From inside to outside, the four rings correspond to the results at iteration 0 (initialization using RoBERTa), 100, 200, 300, respectively.}
    \label{fig:cycle}
\end{figure}

Figure~\ref{fig:cycle} demonstrates \textbf{error reduction}. Samples are indicated by radii of the circle, and classification correctness is indicated by color. For example, if a radius has color orange, blue, blue, blue, then it is mis-classified at iteration 0, and correctly classified at iteration 100, 200, and 300.
We can see that Self-training suffers from error accumulation, as around 2\% more samples are mis-classified between iteration 200 and 300.
In contrast, in {\ours}, a noticeable amount of incorrect predictions are rectified, and the accuracy improves by more than 15\% after 300 iterations. 

%% file: 0-conclusion.tex
\section{Discussion and Conclusion}

In this paper, we propose a differentiable self-training framework, {\ours}, which formulates the teacher-student framework in self-training as a Stackelberg game. 
The formulation treats the student as the leader, and the teacher as the follower. In {\ours}, the student is in an advantageous position by recognizing the follower's strategy. 
In this way, we can find a better descent direction for the student and can stabilize training. 
Empirical results on weakly- and semi-supervised natural language processing tasks suggest the superiority of {\ours} to conventional self-training.

Conventional self-training is a heuristic and does not pose a well-defined optimization problem. In conventional methods, the teacher optimizes an implicit function through different components, e.g., pseudo-labels. We follow this convention and formulate self-training as a Stackelberg game. Our formulation is a principle that can motivate follow-up works.

In our Stackelberg game formulation \eqref{eq:diff-self-train}, the student’s utility function is the objective function of the minimization problem. The teacher’s utility function is an implicit function, which can be written as the following:
\begin{align*}
    \text{Utility}(\text{teacher}) &= \cD(\text{teacher}, \text{student}) \\
    & + \cR_1(\text{teacher}, \text{confidence}) \\
    & + \cR_2(\text{teacher}, \text{uncertainty}).
\end{align*}
Here, the first term $\cD$ is some divergence between the teacher and the student (i.e., the KL-divergence in \eqref{eq:student-loss}), and the two regularizers $\cR_1$ and $\cR_2$ are defined implicitly.
That is, $\cR_1$ regularizes model confidence (realized by the soft pseudo-labels in \eqref{eq:soft-label}), and $\cR_2$ regularizes model uncertainty (realized by the sample weights in \eqref{eq:sample-weight}).
Even though the utility function of the teacher is implicit, the solution of it is explicitly given, namely the teacher’s strategy $F(\theta^T)$ in \eqref{eq:diff-self-train} is the solution to the teacher’s implicit utility.

Because the strategy of the teacher is explicit (in contrast to implicitly defined by an optimization problem), the teacher’s utility is maximized with such a strategy. Thus, equilibrium of the Stackelberg game exists, and every local optimum of the minimization problem is an equilibrium. In this work, we use off-the-shelf algorithms (Adam and AdamW) to find local minima of \eqref{eq:diff-self-train} (a.k.a. equilibria of the game).

We remark that in a Stackelberg game, the names ``leader'' and ``follower'' indicate the relative importance and priority of the two players. In our framework, we use the student model for prediction. Therefore, the student model is more important than the teacher, so that we grant it a higher priority and say it is the leader.

We remark that self-distillation~\citep{furlanello2018born} is a special case of self-training, which is a supervised learning method. We can also differentiate the teacher model in self-distillation, such that {\ours} can be extended to supervised learning.

%% file: 0-acknowledgement.tex
\section*{Acknowledgements}

This work was supported in part by NSF IIS-2008334, IIS-2106961, CAREER IIS-2144338, and ONR MURI N00014-17-1-2656.

%% file: 0-broader-impact.tex
\section*{Ethical Statement}

This paper proposes Differentiable Self-Training (\ours), a self-training framework for NLP tasks.
We demonstrate that the DRIFT framework can be used for text classification and named entity recognition tasks. Moreover, the framework is also demonstrated to be effective for semi-supervised classification on graphs. We use publicly available datasets to conduct all the experiments. And the proposed method is built using public code bases.
We do not find any ethical concerns.

%% file: 0-appendix.tex

\section{Semi-Supervised Learning on Graphs}
\label{app:graph}

\begin{table*}[t!]
\centering \small
\begin{tabular}{l|cccc|cccc|cccc}
\toprule
\textbf{Dataset} & \multicolumn{4}{c|}{\textbf{Cora}} & \multicolumn{4}{c|}{\textbf{Citeseer}} & \multicolumn{4}{c}{\textbf{Pubmed}} \\ \cline{1-1}
\textbf{Labels per class} & 10 & 20 & 50 & 100 & 10 & 20 & 50 & 100 & 10 & 20 & 50 & 100 \\ \midrule
\multicolumn{1}{l}{\textbf{Baselines}} \\
GCN~\cite{kipf2016semi} & 74.5 & 77.4 & 81.6 & 85.1 & 67.1 & 69.5 & 71.9 & 74.9 & 71.0 & 75.1 & 81.8 & 84.8 \\
Self-training~\cite{lee2013pseudo} & 74.4 & 79.1 & 83.5 & 85.1 & 70.5 & 73.1 & 75.1 & 76.2 & 71.8 & 75.2 & 82.5 & 84.6 \\
GraphVAT~\cite{feng2019graph} & 75.2 & 78.6 & 83.1 & 85.3 & 67.6 & 70.5 & 72.6 & 75.8 & 71.8 & 75.5 & 82.1 & 85.0 \\
GraphMix~\cite{verma2019graphmix} & 77.3 & 82.3 & 84.8 & 86.0 & 67.1 & 73.9 & 74.5 & 76.9 & 72.9 & 76.1 & 81.9 & 84.4 \\
GRAND~\cite{feng2020graph} & 76.5 & 84.3 & 86.5 & 87.2 & 62.8 & 73.3 & 75.0 & 77.8 & 77.4 & 78.5 & 83.9 & 86.2 \\ \midrule
\multicolumn{1}{l}{\textbf{Ours}} \\
\ours+GCN & 80.4 & 81.8 & 84.6 & 85.6 & \textbf{74.4} & 75.4 & \textbf{75.9} & 77.4 & 72.8 & 78.1 & 83.3 & 85.3 \\
\ours+GRAND & \textbf{82.1} & \textbf{85.4} & \textbf{87.3} & \textbf{87.9} & 74.1 & \textbf{76.0} & 75.7 & \textbf{78.5} & \textbf{79.2} & \textbf{79.3} & \textbf{85.2} & \textbf{86.8} \\
\bottomrule
\end{tabular}
\caption{Accuracy (in \%) of semi-supervised node classification on graphs. For all the splits of a particular dataset, we use the same development and test sets. We report the mean over ten runs. The best results are shown in bold.}
\label{tb:graph-results}
\end{table*}

\begin{table*}[t!]
\centering
\begin{tabular}{c|c|c|c|c|c|c}
\toprule \bf Dataset & \bf \#Nodes & \bf \#Edges & \bf \#Class & \bf \#Dev & \bf \#Test  & \bf \#Features \\ \midrule
Cora & 2,708 & 5,429 & 7 & 500 & 1,000 & 1,433 \\
Citeseer & 3,327 & 4,732 & 6 & 500 & 1,000 & 3,703 \\
Pubmed & 19,717 & 44,338 & 3 & 500 & 1,000 & 500 \\
\bottomrule
\end{tabular}
\caption{Statistics of datasets used in semi-supervised learning on graphs.}
\label{tb:graph-data}
\end{table*}

\noindent \textbf{Datasets. }
We adopt three citation networks: Cora, Citeseer, and Pubmed~\cite{sen2008collective} as benchmark datasets. Their statistics are summarized in Table~\ref{tb:graph-data}. Similar to semi-supervised text classification tasks, for each dataset, we randomly sample $N \in \{10, 20, 50, 100\}$ data points from each class and annotate them with clean labels, while the other data are treated as unlabeled. We use the same development and test sets for all the splits of a particular dataset.

\vspace{0.1in}
\noindent \textbf{Baselines. } In addition to Self-training, we adopt four graph neural network methods as baselines. Note that Self-training uses GCN as its backbone.

\vspace{0.05in}
\noindent $\diamond$
\textit{GCN}~\cite{kipf2016semi} adopts graph convolutions as an information propagation operator on graphs. The operator smooths label information over the graph, such that labeled nodes acknowledge features of unlabeled ones, and predictions are drawn accordingly.

\vspace{0.05in}
\noindent $\diamond$
\textit{GraphVAT}~\cite{feng2019graph} leverages virtual adversarial training on graphs. The method generates perturbations to each data point, and promotes smooth predictions subject to the perturbations.

\vspace{0.05in}
\noindent $\diamond$
\textit{GraphMix}~\cite{verma2019graphmix} is an interpolation-based regularization method. It uses a manifold mixup approach to learning more discriminative node representations.

\vspace{0.05in}
\noindent $\diamond$
\textit{GRAND}~\cite{feng2020graph} performs data augmentation via a random propagation strategy. It also leverages a consistency regularization to encourage prediction consistency across different augmentations. GRAND uses a multi layer perception (MLP) as its backbone.

\vspace{0.1in}
\noindent \textbf{Settings. }
To demonstrate that differentiable self-training can be effectively combined with different models, we adopt {\ours} to two architectures: GCN, which is a graph convolution-based method; and GRAND, which is a MLP-based method that achieves state-of-the-art performance.

\vspace{0.1in}
\noindent \textbf{Results. }
Experimental results are summarized in Table~\ref{tb:graph-results}.
Notice that Self-training outperforms GCN. This is because while GCN only implicitly uses information of the unlabeled nodes, Self-training directly utilizes such information via the pseudo-labels.
Furthermore, {\ours}+GCN enhances the performance of Self-training.
The other baselines (e.g., GraphVAT, Graphmix, GRAND), which are refinements and substitutions to the graph convolution operation, outperforms vanilla GCN.
By equipping GRAND with differentiable self-training, {\ours}+GRAND achieves the best performance in 10 out of 12 experiments. The performance gain is more pronounced when there are only a few labeled samples, e.g., {\ours}+GRAND improves GRAND by more than 11\% when there are 10 labeled samples per class.

\begin{figure}[htb!]
    \centering
    \begin{subfigure}[t]{0.49\linewidth}
         \centering
         \includegraphics[width=\linewidth]{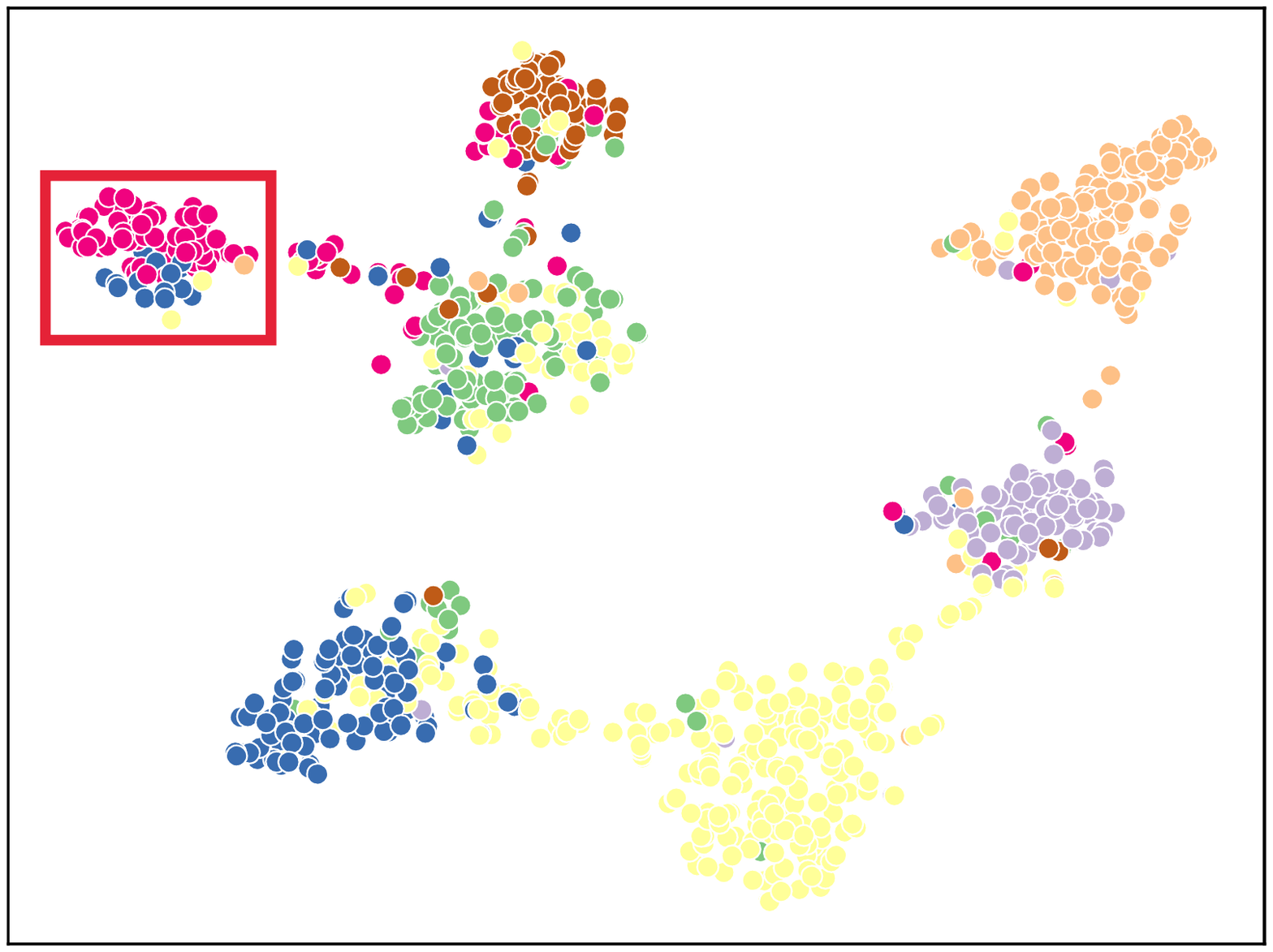}
         \vskip -0.1in
         \caption{Self-training on Cora.}
         \label{fig:tsne:self-cora}
     \end{subfigure} \hfill
     \begin{subfigure}[t]{0.49\linewidth}
         \centering
         \includegraphics[width=\linewidth]{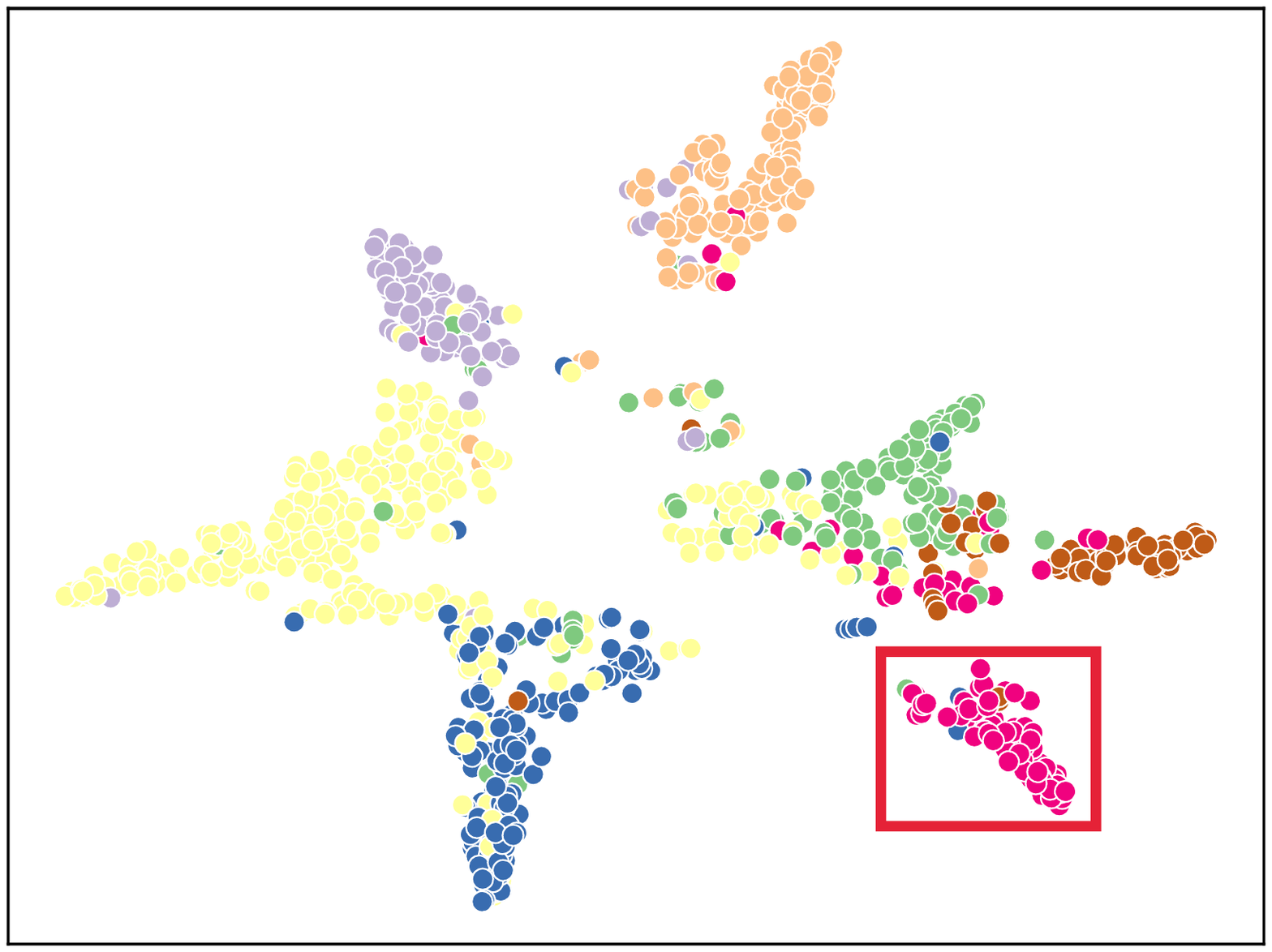}
         \vskip -0.1in
         \caption{{\ours} on Cora.}
         \label{fig:tsne:diff-cora}
     \end{subfigure}
     \begin{subfigure}[t]{0.49\linewidth}
         \centering
         \includegraphics[width=\linewidth]{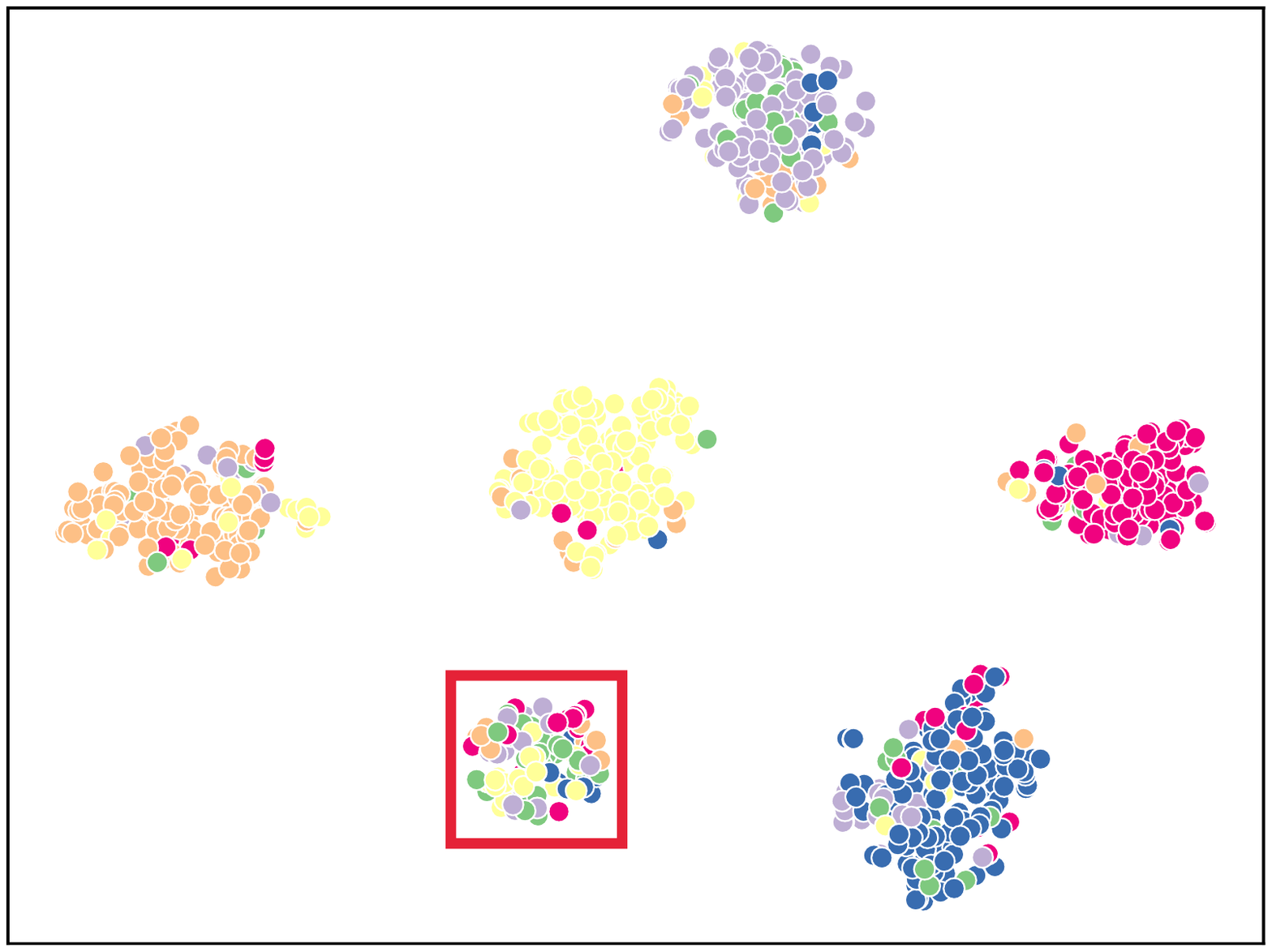}
         \vskip -0.1in
         \caption{Self-training on Citeseer.}
         \label{fig:tsne:self-citeseer}
     \end{subfigure} \hfill
     \begin{subfigure}[t]{0.49\linewidth}
         \centering
         \includegraphics[width=\linewidth]{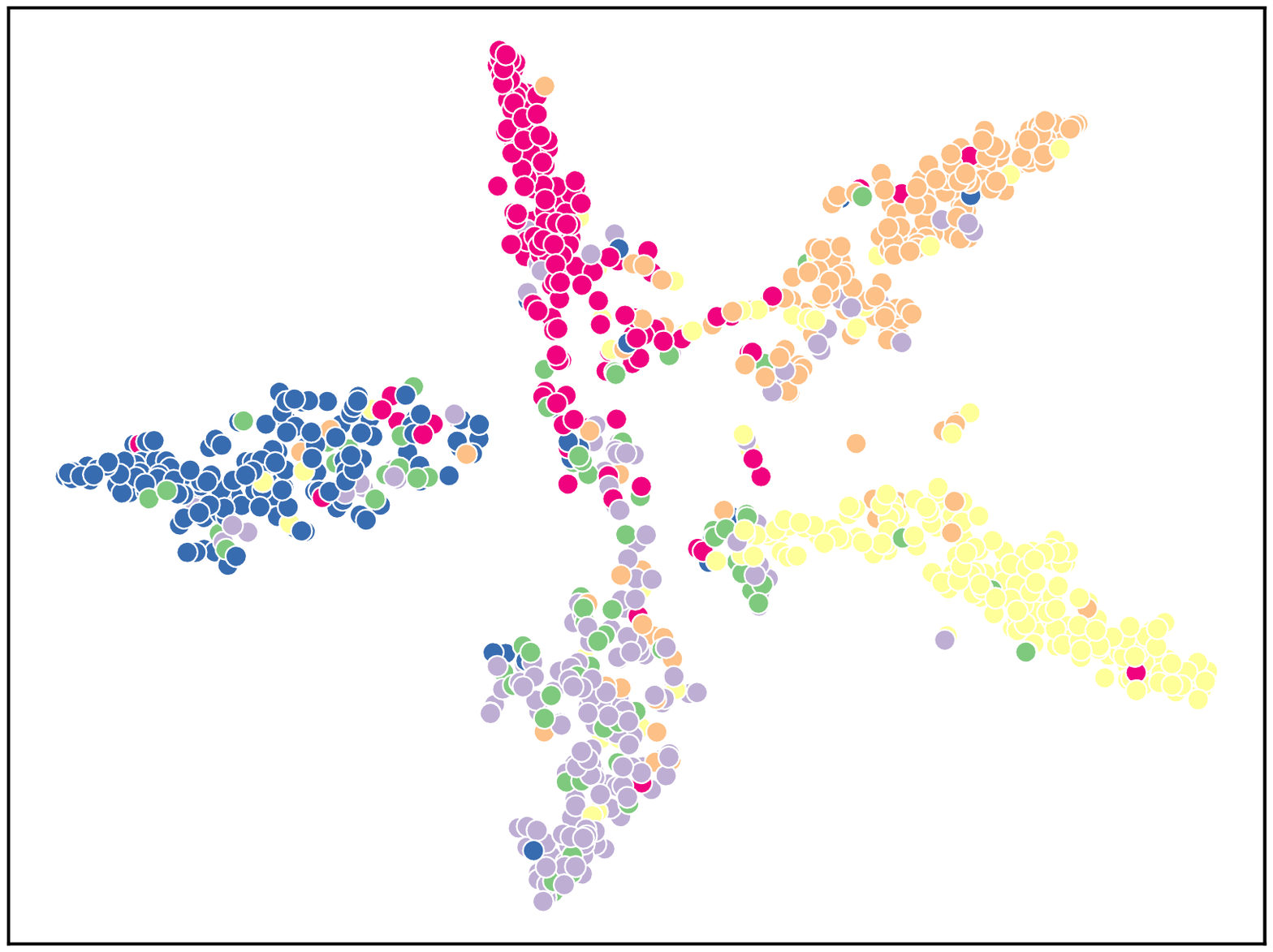}
         \vskip -0.1in
         \caption{{\ours} on Citeseer.}
         \label{fig:tsne:diff-citeseer}
     \end{subfigure}
     \vskip -0.05in
     \caption{t-SNE plots of Self-training and {\ours} on Cora and Citeseer. Each color denotes a different class.}
     \label{fig:tsne}
\end{figure}

\vspace{0.1in}
\noindent
\textbf{Visualization of learned representations. }
Figure~\ref{fig:tsne} visualizes the learned representations of Self-training and {\ours}.
From Fig.~\ref{fig:tsne:self-cora}, we can see that Self-training mixes the representations of the red class and the blue class, as indicated in the red box. Such erroneous classification is alleviated by {\ours} (Fig.~\ref{fig:tsne:diff-cora}).
On Citeseer, notice that Self-training generates a meaningless cluster (Fig.~\ref{fig:tsne:self-citeseer}), which is a sign that Self-training overfits on the label noise.

\section{Classification and Named Entity Recognition Datasets}
\label{app:nlp-dataset}

Dataset statistics for the classification and named entity recognition tasks are presented in Table~\ref{tb:text-dataset}.
\begin{table*}[htb!]
\centering
\begin{tabular}{L|C|C|C|C|C}
\toprule \bf Dataset & \bf Task & \bf \#Class &\bf \#Train & \bf \#Dev & \bf \#Test \\ \midrule
AGNews &Topic &4 &108k & 12k & 7.6k \\ 
IMDB &Sentiment &2 &20k & 2.5k & 2.5k \\
Yelp &Sentiment& 2 & 30.4k & 3.8k & 3.8k \\
Amazon &Sentiment& 2 & 25k & 2.5k & 2.5k \\
MIT-R & Slot Filling &  9& 6.6k & 1.0k & 1.5k \\
CoNLL-03 & NER & 4 & 14.0k &3.2k &3.4k  \\
Webpage & NER & 4&385&99 &135 \\
Wikigold & NER & 4&1.1k&280 &274 \\
BC5CDR & NER & 2& 4.5k &4.5k& 4.7k \\
\bottomrule
\end{tabular}
\caption{Statistics of datasets used in text classification and named entity recognition tasks.}
\label{tb:text-dataset}
\end{table*}

\section{Weak Supervision Sources}
\label{app:dataset}

There are two types of semantic rules that we apply as weak supervisions:
\begin{itemize}
    \item \textit{Keyword Rule}: \texttt{HAS(x, L) $\rightarrow$ C}. If $x$ matches one of the words in the list $L$, we label it as $C$.
    \item \textit{Pattern Rule}: \texttt{MATCH(x, R) $\rightarrow$ C}. If $x$ matches the regular expression $R$, we label it as $C$.
\end{itemize}
Two examples of semantic rules on AGNews and IMDB are given in Table~\ref{tab:agnews} and Table~\ref{tab:imdb}.

All of the weak supervisions, i.e., linguistic rules, are from existing literature.
The details are listed below:
\begin{itemize}
    \item \textit{AGNews, IMDB, Yelp}: We use the rules in~\citet{ren2020denoise}.
    \item \textit{MIT-R}:  We use the rules in~\citet{Awasthi2020Learning}.
    \item \textit{CoNLL-03, WebPage, Wikigold}: We use the keywords in~\citet{liang2020bond}.
    \item \textit{BC5CDR}: We use the keywords in~\citet{shang2018learning}. Note that for simplicity, we do not use AutoPhrase to extract external keywords. Such an approach requires external corpus and extra parameter-tuning.
\end{itemize}

\begin{table*}[htb!]
\small \centering
\begin{tabular}
{p{400pt} }
\toprule
\textbf{Rule} \\
\midrule
\texttt{[war, prime minister, president, commander, minister,  military, militant, kill, operator] $\rightarrow$ POLITICS} \\ \hline
\texttt{[baseball, basketball, soccer, football, boxing,  swimming, world cup, nba,olympics,final, fifa] $\rightarrow$ SPORTS} \\ \hline
\texttt{[delta, cola, toyota, costco, gucci, citibank, airlines] $\rightarrow$ BUSINESS} \\ \hline
\texttt{[technology, engineering, science, research, cpu, windows, unix, system, computing,  compute] $\rightarrow$ TECHNOLOGY } \\
\bottomrule
\end{tabular}
\caption{Examples of semantic rules on AGNews.}
\label{tab:agnews}
\end{table*}

\begin{table*}[htb!]
\small \centering
\begin{tabular}
{p{400pt} }
\toprule
\textbf{Rule} \\
\midrule
\texttt{[masterpiece, outstanding, perfect, great, good, nice, best, excellent, worthy, awesome, enjoy, positive, pleasant, wonderful, amazing, superb, fantastic, marvellous, fabulous] $\rightarrow$ POS} \\ \hline
\texttt{[bad, worst, horrible, awful, terrible, crap, shit, garbage, rubbish, waste] $\rightarrow$ NEG} \\ \hline
\texttt{[beautiful, handsome, talented]$\rightarrow$ POS} \\ \hline
\texttt{ [fast forward, n t finish] $\rightarrow$ NEG}  \\ \hline
\texttt{[well written, absorbing,attractive, innovative, instructive,interesting, touching, moving]$\rightarrow$ POS} \\ \hline
\texttt{[to sleep, fell asleep, boring, dull, plain]$\rightarrow$ NEG}  \\ \hline
\texttt{[ than this,  than the film,  than the movie]$\rightarrow$ NEG}  \\ \hline
\texttt{MATCH(x, *PRE*EXP* ) $\rightarrow$ POS} 
\texttt{PRE} = [will ,  ll , would , can't wait to ]
\texttt{EXP} = [ next time,  again,  rewatch,  anymore,  rewind] \\ \hline
\texttt{PRE} = [highly , do , would , definitely , certainly , strongly , i , we ] 
\texttt{EXP} = [ recommend,  nominate] \\ \hline
\texttt{PRE} = [high , timeless , priceless , has , great , of , real , instructive ]
\texttt{EXP} = [ value,  quality,  meaning,  significance] \\ 
\bottomrule
\end{tabular}
\caption{Examples of semantic rules on IMDB.}
\label{tab:imdb}
\end{table*}

\section{Training Details}
\label{app:training}

We use a validation set to tune {\ours} as well as all the baseline methods. We report the test result of the best model on the validation set. All the experimental results have passed a paired t-test with $p<0.05$.

\subsection{Baseline Settings}
We implement the GraphVAT method by ourselves.
For the other baselines, we follow the official release: \\
(1) MixText: \url{https://github.com/GT-SALT/MixText/}; \\
(2) BOND: \url{https://github.com/cliang1453/BOND}; \\
(3) UAST: \url{https://github.com/microsoft/UST}; \\
(4) WeSTClass: \url{https://github.com/yumeng5/WeSTClass}; \\
(5) GCN: \url{https://github.com/tkipf/pygcn}; \\
(6) GRAND: \url{https://github.com/THUDM/GRAND}; \\
(7) GraphMix: \url{https://github.com/vikasverma1077/GraphMix}.

\begin{table*}[htb!]
\centering \small
\begin{tabular}{c|c|c|c|c|c|c|c|c}
\toprule 
\bf Hyper-parameter &\bf AGNews& \bf  IMDB & \bf Yelp  & \bf MIT-R & \bf CoNLL-03 & \bf Webpage & \bf Wikigold & \bf BC5CDR \\ \midrule 
Dropout Ratio & \multicolumn{8}{@{\hskip1pt}c@{\hskip1pt}}{0.1}  \\ \hline
Maximum Tokens  & 128 & 256 & 512 & 64 & 128 & 128 & 128 & 128 \\ \hline 
Batch Size  & 32 & 16 & 16 & 64& 32& 32& 32& 32 \\ \hline
Weight Decay & \multicolumn{8}{@{\hskip1pt}c@{\hskip1pt}}{$10^{-4}$}  \\ \hline
Learning Rate &\multicolumn{8}{@{\hskip1pt}c@{\hskip1pt}}{$10^{-5}$}   \\ \hline
Initialization Steps & 160 & 160 & 200 & 150 & 900 & 300 & 3500 & 1500 \\ \hline
$T$ &  3000 & 2500 & 2500 & 1000& 1800& 200& 700& 1000 \\ \hline
$\alpha$ & 0.95 & 0.9 & 0.95 & 0.9 & 0.9 & 0.95 & 0.9 & 0.9 \\ \hline
$\tau$ & 0.5 & 0.5  & 0.5  & 0.5  & 0.5  & 0.5  & 0.5  & 0.5   \\ 
\bottomrule
\end{tabular}
\caption{Hyper-parameter configurations for weakly-supervised text classification.}
\label{tab:hyperparameter-weak}
\end{table*}

\begin{table*}[htb!]
\centering
\begin{tabular}{c|c|c|c}
\toprule 
\bf Hyper-parameter &\bf AGNews& \bf  IMDB & \bf Amazon  \\ \midrule 
Dropout Ratio & \multicolumn{3}{@{\hskip1pt}c@{\hskip1pt}}{0.1}  \\ \hline
Maximum Tokens  & 128 & 256 & 256 \\ \hline 
Batch Size  & 32 & 16 & 16 \\ \hline
Weight Decay & \multicolumn{3}{@{\hskip1pt}c@{\hskip1pt}}{$10^{-4}$}  \\ \hline
Learning Rate &\multicolumn{3}{@{\hskip1pt}c@{\hskip1pt}}{$10^{-5}$}   \\ \hline
Initialization Steps & 1200 & 1000 & 800  \\ \hline
$T$ &  4000 & 3000 & 4000  \\ \hline
$\alpha$ & 0.95 & 0.99 & 0.9  \\ \hline
$\tau$ & 0.6 & 0.5  & 0.5    \\ 
\bottomrule
\end{tabular}
\caption{Hyper-parameter configurations for semi-supervised text classification.}
\label{tab:hyperparameter-semi}
\end{table*}

\subsection{Weakly-Supervised Text Classification}

Hyper-parameters are shown in Table~\ref{tab:hyperparameter-weak}.

\subsection{Semi-Supervised Text Classification}

We implement TextCNN with Pytorch~\cite{paszke2019pytorch}.
We use the pre-trained 300 dimension FastText embeddings\footnote{We use the 1 million word vectors trained on Wikipedia 2017, UMBC webbase corpus and news dataset, which is available online: \url{https://fasttext.cc/docs/en/english-vectors.html}.} as the input vectors.
Then, we set the 
filter window sizes to 2, 3, 4, 5 with 500 feature maps each. We train the model for 100 iterations as initialization, and set $T=1000$ during self-training. We use Stochastic Gradient Descent (SGD) with momentum $m=0.9$ and we set the learning rate to $5 \times 10^{-4}$. We set the dropout rate to 0.5 for the linear layers after the CNN. We tune the weight decay in $\left[10^{-4}, 10^{-5}, 10^{-6}, 10^{-7}\right]$.

Hyper-parameters are shown in Table~\ref{tab:hyperparameter-semi}.

\subsection{Semi-Supervised Learning on Graphs}
Our method serves as an efficient drop-in module to existing methods. There are only two parameters that we tune in the experiments: the exponential moving average rate $\alpha$ and the temperature $\tau$ of the soft pseudo-labels. For all the three datasets, we set $\alpha=0.99$. For the temperature parameter, we use the following settings.
\begin{itemize}
    \item Cora: $1/\tau=3.0$ for GRAND and $1/\tau=4.0$ for GCN.
    \item Citeseer: $1/\tau=3.0$ for GRAND and $1/\tau=3.5$ for GCN.
    \item Pubmed: $1/\tau=3.0$ for GRAND and $1/\tau=4.0$ for GCN.
\end{itemize}
Other hyper-parameters and tricks used in training follow the corresponding works.